\documentclass[10pt,twocolumn,letterpaper]{article}

\usepackage[pagenumbers]{cvpr}
\usepackage{array}

\usepackage{graphicx}
\usepackage{amsmath}
\usepackage{amssymb}
\usepackage{booktabs}
 
\usepackage{xcolor}
\usepackage{colortbl}
\usepackage{times}
\usepackage{epsfig}
\usepackage{soul}
\usepackage{caption}
\usepackage{subcaption}
\usepackage{bbold}
\usepackage[mathletters]{ucs}
\usepackage[utf8x]{inputenc}
\renewcommand{\paragraph}[1]{\vspace{1.25mm}\noindent\textbf{#1}}
\makeatletter\renewcommand\paragraph{\@startsection{paragraph}{4}{\z@} {.5em \@plus1ex \@minus.2ex}{-.5em}{\normalfont\normalsize\bfseries}}\makeatother
\usepackage{xfp}

\definecolor{sasacolor}{HTML}{000000}
\definecolor{vitcolor}{HTML}{fc8e62}
\definecolor{convcolor}{HTML}{412F8A}
\definecolor{swincolor}{HTML}{fc6562}
\definecolor{pa}{HTML}{C209C1}

\definecolor{citationcolor}{HTML}{007ED2}
\newcommand{\sasacolor}[1]{\textcolor{sasacolor}{#1}}
\newcommand{\vitcolor}[1]{\textcolor{vitcolor}{#1}}
\newcommand{\convcolor}[1]{\textcolor{convcolor}{#1}}
\newcommand{\swincolor}[1]{\textcolor{swincolor}{#1}}
\newcommand{\natcolor}[1]{\textcolor{pa}{#1}}

\usepackage[
    pagebackref,
    breaklinks,
    colorlinks,
    citecolor=citationcolor,
    linkcolor=blue
]{hyperref}

\usepackage[capitalize]{cleveref}
\crefname{section}{Sec.}{Secs.}
\Crefname{section}{Section}{Sections}
\Crefname{table}{Table}{Tables}
\crefname{table}{Tab.}{Tabs.}
\Crefname{appendix}{Appendix}{Appendixes}

\makeatletter
\newcommand{\settitle}{\@maketitle}
\makeatother

\def\cvprPaperID{xxxxx} 
\def\confName{CVPR}
\def\confYear{2023}

\newcommand{\vb}{\vitcolor{$\mathbf{\circ}$\,}} 
\newcommand{\wb}{\swincolor{$\mathbf{\circ}$\,}} 
\newcommand{\cb}{\convcolor{$\bullet$\,}} 
\newcommand{\nb}{\natcolor{$\mathbf{\circ}$\,}}
\newcommand{\grayrow}{\rowcolor[gray]{.95}}
\newcommand{\ours}{\grayrow\nb}
\newcommand{\sab}{\grayrow\sasacolor{$\mathbf{\circ}$\,}} 

\DeclareMathAlphabet\mathbfcal{OMS}{cmsy}{b}{n}
\newcommand{\natten}{$\mathcal{N}ATTEN$}
\newcommand{\nattenb}{$\mathbfcal{N}\mathbf{ATTEN}$}

\newcommand{\ips}{\tiny{imgs/sec}}
\newcommand{\fps}{\tiny{fps}}

\newcommand{\salientscheme}{light}

\begin{document}
\title{Neighborhood Attention Transformer}

\author{Ali Hassani\textsuperscript{1}, Steven Walton\textsuperscript{1}, Jiachen Li\textsuperscript{1}, Shen Li\textsuperscript{3}, Humphrey Shi\textsuperscript{1,2} \\
{\small \textsuperscript{1}SHI Labs @ U of Oregon \& UIUC, \textsuperscript{2}Picsart AI Research (PAIR), \textsuperscript{3}Meta/Facebook AI}\\
{\small \textbf{\url{https://github.com/SHI-Labs/Neighborhood-Attention-Transformer}}}
}
\twocolumn[{
\settitle
\begin{center}
\vspace{-2mm}
    \includegraphics[width=0.99\textwidth]{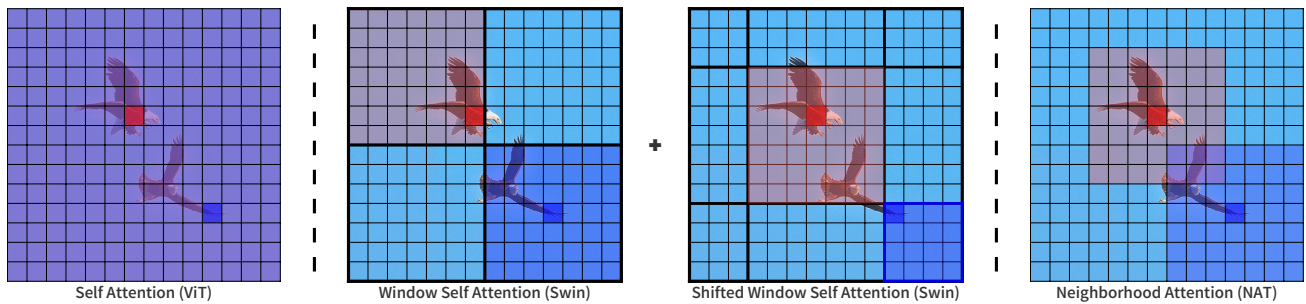}
    \captionsetup{type=figure}
    \captionof{figure}{
    \textbf{An illustration of attention spans in Self Attention, (Shifted) Window Self Attention, and our Neighborhood Attention.}
    Self Attention allows each token to attend to everything. 
    Window Self Attention divides self attention into non-overlapping sub-windows, and is followed by Shifted Window Self Attention, which allows for out-of-window interactions that are necessary to receptive field expansion. 
    Neighborhood Attention localizes attention to a neighborhood around each token, introducing local inductive biases, maintaining translational equivariance, and allowing receptive field growth without needing extra operations.
    }
    \label{fig:teaser}
\end{center}
}]

\thispagestyle{empty}

\begin{abstract}
We present \textbf{Neighborhood Attention (NA)}, the first efficient and scalable sliding window attention mechanism for vision.
NA is a pixel-wise operation, localizing self attention (SA) to the nearest neighboring pixels, and therefore enjoys a linear time and space complexity compared to the quadratic complexity of SA.
The sliding window pattern allows NA's receptive field to grow without needing extra pixel shifts, and preserves translational equivariance, unlike Swin Transformer's Window Self Attention (WSA).
We develop \textbf{\natten} (Neighborhood Attention Extension), a Python package with efficient C++ and CUDA kernels, which allows NA to run up to 40\% faster than Swin's WSA while using up to 25\% less memory.
We further present \textbf{Neighborhood Attention Transformer (NAT)}, a new hierarchical transformer design based on NA that boosts image classification and downstream vision performance.
Experimental results on NAT are competitive; NAT-Tiny reaches 83.2\% top-1 accuracy on ImageNet, 51.4\% mAP on MS-COCO and 48.4\% mIoU on ADE20K, which is 1.9\% ImageNet accuracy, 1.0\% COCO mAP, and 2.6\% ADE20K mIoU improvement over a Swin model with similar size.
To support more research based on sliding window attention, we open source our project and release our checkpoints.
\end{abstract}

\vspace{-3mm}
\section{Introduction}
\label{sec:intro}
Convolutional neural networks (CNNs)~\cite{lecun1989backpropagation} have been the de facto standard architecture for computer vision models across different applications for years. 
AlexNet~\cite{krizhevsky2012imagenet} showed their usefulness on ImageNet~\cite{deng2009imagenet}, and many others followed suit with architectures such as VGG~\cite{simonyan2014very}, ResNet~\cite{he2016deep}, and EfficientNet~\cite{tan2019efficientnet}. 
Transformers~\cite{vaswani2017attention} on the other hand, were originally proposed as attention-based models for natural language processing (NLP), trying to exploit the sequential structure of language. 
They were the basis upon which BERT~\cite{devlin2019bert} and GPT~\cite{radford2018improving,radford2019language,brown2020language} were built, and they continue to be the state of the art architecture in NLP.

In late 2020, Vision Transformer (ViT)~\cite{dosovitskiy2020image} was proposed as an image classifier using only a Transformer Encoder operating on an embedded space of image patches, mostly for large-scale training. A number of other methods followed, attempting to increase data efficiency~\cite{touvron2021training,el2021xcit,hassani2021escaping}, eventually making such Transformer-like models the state of the art in ImageNet-1K classification (without pre-training on large-scale datasets such as JFT-300M).

These high-performing Transformer-like methods are all based on Self Attention (SA), the basic building block in the original Transformer~\cite{vaswani2017attention}. SA has a linear complexity with respect to the embedding dimension (excluding linear projections), but a quadratic complexity with respect to the number of tokens.
In the scope of vision, the number of tokens is typically in linear correlation with image resolution.
As a result, higher image resolution results in a quadratic increase in complexity and memory usage in models strictly using SA, such as ViT.
The quadratic complexity has prevented such models from being easily applicable to downstream vision tasks, such as object detection and segmentation, in which image resolutions are usually much larger than classification. 
Another problem is that convolutions benefit from inductive biases such as locality, and the 2-dimensional spatial structure, while dot-product self attention is a global 1-dimensional operation by definition. 
This means that some of those inductive biases have to be learned with either large sums of data~\cite{dosovitskiy2020image} or advanced training techniques and augmentations~\cite{touvron2021training,hassani2021escaping}.

Local attention modules were therefore proposed to alleviate these issues. 
Stand-Alone Self-Attention (SASA)~\cite{ramachandran2019stand} was one of the earliest applications of local window-based attention to vision, where each pixel attends to a window around it.
Its explicit sliding window pattern is identical to that of \textit{same} convolutions, with zero paddings around and a simple 2-dimensional raster scan, therefore maintaining translational equivariance.
SASA was aimed at replacing convolutions in a ResNet, and was shown to have a noticeable improvement over baselines.
However, the authors noted SASA was limited in terms of speed due to the lack of an efficient implementation similar to that of convolutions.
Swin~\cite{liu2021swin} on the other hand was one of the first hierarchical vision transformers based on local self attention. 
Its design and the shifted-window self attention allowed it to be easily applicable to downstream tasks, as they made it computationally feasible, while also boosting performance through the additional biases injected. Swin's localized attention, however, first applies self attention to non-overlapping windows and then shifts the windows, the motivation of which was sliding window methods such as SASA suffering throughput bottlenecks.
HaloNet~\cite{vaswani2021scaling} used a haloing mechanism that localizes self attention for blocks of pixels at a time, as opposed to pixel-wise. One of their key motivations for this was also noted to be the lack of an efficient sliding window attention.

\begin{figure}[t]
    \centering
    \includegraphics[width=0.475\textwidth]{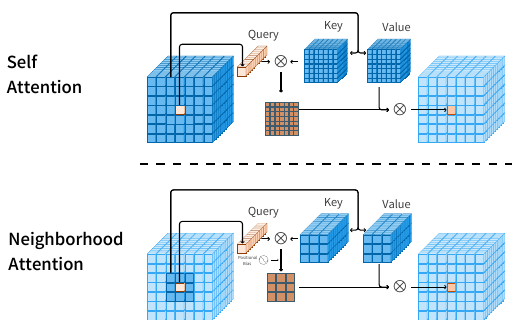}
    \caption{
    \textbf{Illustration of the query-key-value structure of Neighborhood Attention (NA) vs Self Attention (SA) for a single pixel. }
    SA allows each pixel to attend to every other pixel, whereas NA localizes attention for each pixel to a neighborhood around itself. 
    Therefore, each pixel's attention span is usually different from the next.
    }
    \label{fig:attn}
\end{figure}

In this work, we revisit explicit sliding window attention mechanisms, and propose Neighborhood Attention (NA). 
NA localizes SA to each pixel's nearest neighbors, which is not necessarily a fixed window around the pixel.
This change in definition allows all pixels to maintain an identical attention span, which would otherwise be reduced for corner pixels in zero-padded alternatives (SASA).
NA also approaches SA as its neighborhood size grows, and is equivalent to SA at maximum neighborhood.
Additionally, NA has the added advantage of maintaining translational equivariance~\cite{vaswani2021scaling}, unlike blocked and window self attention.
We develop \natten{}, a Python package with efficient C++ and CUDA kernels that allow NA to run even faster than Swin's WSA in practice, while using less memory.
We build Neighborhood Attention Transformer (NAT), which achieves competitive results across vision tasks. 

To summarize, our main contributions are:
\begin{enumerate}
    \item Proposing \textbf{Neighborhood Attention (NA)}: A simple and flexible explicit sliding window attention mechanism that localizes each pixel's attention span to its nearest neighborhood, approaches self attention as its span grows, and maintains translational equivariance. 
    We compare NA in terms of complexity and memory usage to self attention, window self attention, and convolutions.
    
    \item Developing efficient C++ and CUDA kernels for NA, including the \textbf{tiled NA} algorithm, which allow NA to run up to 40\% faster than Swin's WSA while using up to 25\% less memory.
    We release them under a new Python package for explicit sliding window attention mechanisms, \nattenb, 
    to provide easy-to-use modules with autograd support that can be plugged into any existing PyTorch pipeline.
    
    \item Introducing \textbf{Neighborhood Attention Transformer (NAT)}, a new efficient, accurate, and scalable hierarchical transformer based on NA. 
    We demonstrate its effectiveness on both classification and downstream tasks. 
    For instance, NAT-Tiny reaches 83.2\% top-1 accuracy on ImageNet with only 4.3 GFLOPs and 28M parameters, and 51.4\% box mAP on MS-COCO and 48.4\% mIoU on ADE20K, significantly outperforming both Swin Transformer and ConvNeXt~\cite{liu2022convnet}.
\end{enumerate}

\begin{figure}[!t]
    \centering
    \includegraphics[width=0.49\textwidth]{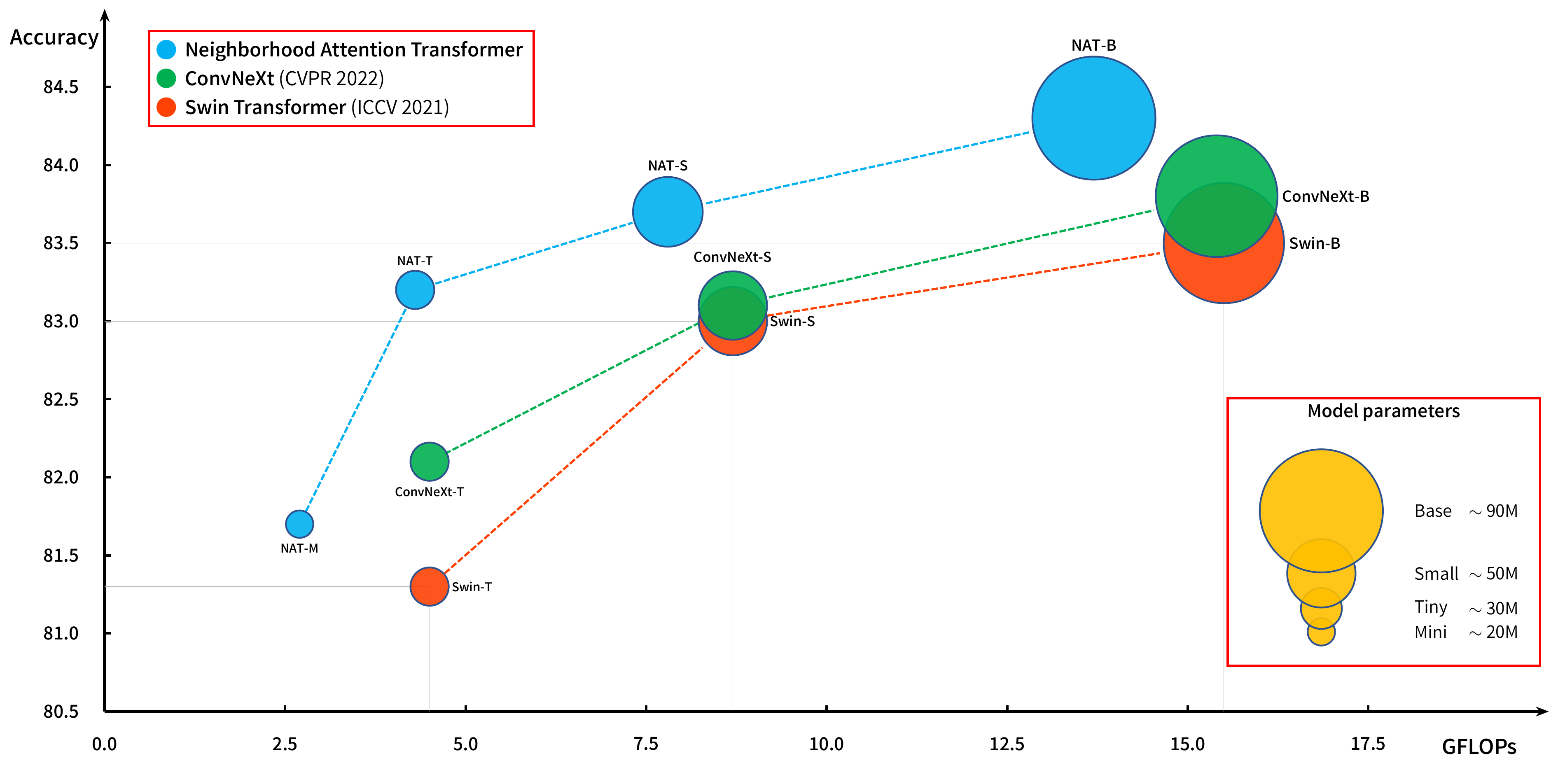}
    \caption{
    \textbf{ImageNet-1K classification performance versus compute, with bubble size representing the number of parameters.} 
    NAT outperfoms both Swin Transformer and ConvNeXt in classification with fewer FLOPs, and a similar number of parameters.
    }
    \label{fig:natperformance}
\end{figure}
\section{Related Works}
\label{sec:related}
\looseness=-1 In this section, we briefly review the original Self Attention (SA)~\cite{vaswani2017attention}, some of the notable vision transformers and Transformer-like architectures \cite{dosovitskiy2020image,touvron2021training}, some of the notable local attention-based vision transformers \cite{liu2021swin,vaswani2021scaling}, and a recent CNN which provides an up-to-date baseline for attention-based models.

\subsection{Self Attention}
Scaled dot-product attention was defined by Vaswani et al.~\cite{vaswani2017attention} as an operation on a query and a set of key-value pairs. The dot product of query and key is computed and scaled. Softmax is applied to the output in order to normalize attention weights, and is then applied to the values. It can be expressed as follows:
\begin{equation}
    Attention ( Q, K, V ) = softmax \left( \frac{Q K^T}{\sqrt{d}} \right) V,
    \label{eq:attention}
\end{equation}
where $d$ is embedding dimension.
Self attention applies dot-product attention over linear projections of the same input as both the query and key-value pairs.
In Transformers, the multi-headed variants of attention and self attention are typically applied.
Multi-headed attention applies dot-product attention multiple times over different embeddings, hence forming attention heads. 
Given an input $X \in \mathbb{R}^{n \times d}$, where $n$ is the number of tokens and $d$ is the embedding dimension, this operation has a complexity of $\mathcal{O}(n^2 d)$ and a space complexity of $\mathcal{O}(n^2)$ for the attention weights.

\subsection{Vision Transformer}
Dosovitskiy et al.~\cite{dosovitskiy2020image} proposed a Transformer-based image classifier that merely consists of a Transformer encoder~\cite{vaswani2017attention} and an image tokenizer, named \textbf{Vi}sion \textbf{T}ransformer (\textbf{ViT}). 
Previous works, such as DETR~\cite{carion2020end}, explored CNN-Transformer hybrids for object detection. 
ViT on the other hand proposed a model that would only rely on a single non-overlapping convolutional layer (patching and embedding). 
ViT was pre-trained primarily on the private JFT-300M dataset, and was shown to outperform state-of-the-art CNNs on many benchmarks. 
However, it was also added that when ViT is pre-trained on medium-scale datasets, such as ImageNet-1K and ImageNet-21K, it no longer achieves competitive results.
This was attributed to the lack of inductive biases that are inherent to CNNs, which the authors argued is trumped by large-scale training. 
While this effectively proved ViT inferior in medium-scale training, it provided empirical evidence that Transformer-based models outperform CNNs in larger scales.
ViT paved the way for many more vision transformers, and attention-based models in general, that followed and transferred it to medium-scale learning~\cite{touvron2021training}, and even small-scale learning on much smaller datasets~\cite{hassani2021escaping}.
Touvron et al.~\cite{touvron2021training} extended the study of Vision Transformers by exploring data efficiency. Their \textbf{D}ata-\textbf{e}fficient \textbf{i}mage \textbf{T}ransformer (\textbf{DeiT}) model pushed ViT ahead with minimal architectural changes, and through the use of advanced augmentations and training techniques. 
Their efforts highlighted the true potential of a Transformer-based image classifier in medium-sized data regimes, and inspired many more to adopt their training techniques~\cite{liu2021swin,touvron2021going}.

\subsection{Local Attention}
\textbf{Stand Alone Self Attention (SASA)}~\cite{ramachandran2019stand}, is one of the earliest sliding window self attention patterns, aimed to replace convolutions in existing CNNs. 
It operates similarly to a convolution with zero padding, and extracts key-value pairs by striding the feature map. 
The authors reported a noticeable accuracy improvement, but observed that the implementation suffered high latency despite the lower theoretical cost.
This attention pattern was also adopted in language processing in Longformer~\cite{beltagy2020longformer} (sliding window attention),
and later adopted in Vision Longformer (ViL)~\cite{zhang2021multi}. 
While Longformer and ViL's implementations were different from SASA, they were still not able to scale to larger windows and models as a result of both computational overhead.
Additionally, the reduced receptive field in corner cases caused by padding was not addressed.
Window and \textbf{S}hifted \textbf{Win}dow (\textbf{Swin}) Attention~\cite{liu2021swin} were introduced by Liu et al. as non-sliding window-based self attention mechanisms that partition feature maps and apply self attention to each partition separately. 
This operation has a similar theoretical complexity to SASA, but it can be easily parallelized through batched matrix multiplication.
The shifted variant follows the regular, and as the name suggests shifts the partitioning to allow out-of-window interactions, which are necessary for receptive field growth.
Their proposed model, \textbf{Swin Transformer}, is one of the earliest hierarchical vision transformers. 
It produces pyramid-like feature maps, reducing spatial dimensionality while increasing depth. 
This structure has been commonly used in CNNs over the years, and is why Swin can be easily integrated with other networks for application to downstream tasks, such as detection and segmentation. 
Swin outperformed DeiT, which uses a convolutional teacher, at ImageNet-1K classification. 
Moreover, Swin Transformer became the state-of-the-art method in object detection on MS-COCO and in semantic segmentation on ADE20K.
Vaswani et al.~\cite{vaswani2021scaling} proposed \textbf{HaloNet}, which aimed to avoid SASA's speed issue by replacing it with a new blocked attention pattern.
They noted that while this change relaxes translational equivariance, it can provide a reasonable trade-off with speed and memory.
HaloNet's attention mechanism consists of 3 stages: blocking, haloing, and attention. Input feature maps are blocked into non-overlapping subsets, which will serve as queries. 
Followed by that, ``haloed'' neighboring blocks are extracted, which will serve as keys and values. 
Attention is then applied to the extracted queries and key-value pairs. 
HaloNet was shown to be effective at both reducing cost (compared to SASA) and improving performance, especially when used in conjunction with convolutional layers in the network. 
Many works followed Swin in adopting WSA, such as RegionViT~\cite{chen2022regionvit}, in which a regional token is inserted into every local self attention layer for the purpose of introducing global context. 
This work and HaloNet highlight that the research community has lost interest in sliding window attention patterns in part because they are thought to be inefficient.
We aim to change that by introducing \natten{}.

\subsection{New Convolutional Baselines}
Liu et al.~\cite{liu2022convnet} proposed a new CNN architecture influenced by models such as Swin, dubbed ConvNeXt. These models are not attention-based, and manage to outperform Swin across different vision tasks. This work has since served as a new CNN baseline for fair comparison of convolutional models and attention-based models.

We propose Neighborhood Attention, which by design localizes the receptive field to a window around each query, and therefore would not require additional techniques such as the cyclic shift used by Swin. 
Additionally, Neighborhood Attention maintains translational equivariance, which is traded off for efficiency in methods such as HaloNet and Swin.
We demonstrate that Neighborhood Attention can run even faster than methods such as Swin, while using less memory, with our \natten{} python package.
We introduce a hierarchical transformer-like model with this attention mechanism, dubbed Neighborhood Attention Transformer, and demonstrate its performance compared to Swin on image classification, object detection, and semantic segmentation.
\section{Method}

In this section, we introduce Neighborhood Attention, a localization of self attention (see \cref{eq:attention}) considering the structure of visual data. This not only reduces computational cost compared to self attention, but also introduces local inductive biases, similar to that of convolutions. 
We show that NA is better alternative to the previously proposed SASA~\cite{ramachandran2019stand} in terms of restricting self attention, while being equivalent in theoretical cost.
We then introduce our Python package, \natten{}, which provides efficient implementations of NA for both CPU and GPU acceleration. We discuss the novelties in the extension and how it manages to exceed the speed of Swin's WSA and SWSA, while using less memory.
We finally introduce our model, \textbf{N}eighborhood \textbf{A}ttention \textbf{T}ransformer (\textbf{NAT}), which uses this new mechanism instead of self attention. 
In addition, NAT utilizes a multi-level hierarchical design, similar to Swin~\cite{liu2021swin}, meaning that feature maps are downsampled between levels, as opposed to all at once. 
Unlike Swin, NAT uses overlapping convolutions to downsample feature maps, as opposed to non-overlapping (patched) ones, which have been shown to improve model performance by introducing useful inductive biases~\cite{wu2021cvt,hassani2021escaping}.

\subsection{Neighborhood Attention}
Swin's WSA can be considered one of the fastest existing methods to restrict self attention for the purpose of cutting down the quadratic attention cost.
It simply partitions inputs and applies self attention to each partition separately.
WSA requires to be paired with the shifted variant, SWSA, which shifts those partition lines to allow out-of-window interactions.
This is crucial to expanding its receptive field.
Nevertheless, the most direct way to restrict self attention locally, is to allow each pixel to attend to its neighboring pixels. 
This results in most pixels having a dynamically-shifted window around them, which expands receptive field, and would therefore not need a manual shifted variant. 
Additionally, different from Swin and similar to convolutions, such dynamic forms of restricted self attention can preserve translational equivariance~\cite{vaswani2021scaling} (we analyze translational equivariance in different methods including our own in \cref{appdx:te}.)
Inspired by this, we introduce \textbf{Neighborhood Attention (NA)}.
Given an input $X \in \mathbb{R}^{n \times d}$, which is a matrix whose rows are $d$-dimensional token vectors,
and $X$'s linear projections, $Q$, $K$, and $V$, and relative positional biases $B(i, j)$,
we define attention weights for the $i$-th input with neighborhood size $k$, $\mathbf{A}_{i}^{k}$, as the dot product of the $i$-th input's query projection, and its $k$ nearest neighboring key projections:
\begin{equation}
    \mathbf{A}_{i}^{k} = \begin{bmatrix}Q_i K_{\rho_{1}{(i)}}^T + B_{(i,\rho_{1}{(i)})} \\Q_i K_{\rho_{2}{(i)}}^T + B_{(i,\rho_{2}{(i)})} \\ \vdots \\ Q_i K_{\rho_{k}{(i)}}^T + B_{(i,\rho_{k}{(i)})}\end{bmatrix},
    \label{eq:nattenq}
\end{equation}
where $\rho_{j}{(i)}$ denotes $i$'s $j$-th nearest neighbor.
We then define neighboring values, $\mathbf{V}_{i}^{k}$, as a matrix whose rows are the $i$-th input's $k$ nearest neighboring value projections:
\begin{equation}
    \mathbf{V}_{i}^{k} = \begin{bmatrix}V_{\rho_{1}{(i)}}^T & V_{\rho_{2}{(i)}}^T & \hdots & V_{\rho_{k}{(i)}}^T\end{bmatrix}^T .
    \label{eq:nattenv}
\end{equation}
Neighborhood Attention for the $i$-th token with neighborhood size $k$ is then defined as:
\begin{equation}
    \text{NA}_{k}{(i)} = softmax\left(\frac{\mathbf{A}_{i}^{k}}{\sqrt{d}}\right) \mathbf{V}_{i}^{k},
    \label{eq:natten}
\end{equation}
where $\sqrt{d}$ is the scaling parameter.
This operation is repeated for every pixel in the feature map.
Illustrations of this operation are presented in \cref{fig:attn,appfig:fullneighborhood}.

From this definition, it is easy to see that as $k$ grows, $\mathbf{A}_{i}^{k}$ approaches self attention weights, and $\mathbf{V}_{i}^{k}$ approaches $V_i$ itself, therefore Neighborhood Attention approaches self attention.
This is the key difference between NA and SASA~\cite{ramachandran2019stand}, where each pixel attends to a window around it with padding around the input to handle edge cases.
It is thanks to this difference that NA approaches self attention as window size grows, which does not hold true in SASA, due to the zero padding around the input.

\subsection{Tiled NA and \texorpdfstring{\nattenb}{NATTEN}}

Restricting self attention in a pixel-wise manner has not been well-explored in the past, primarily because it was considered a costly operation~\cite{ramachandran2019stand,liu2021swin,vaswani2021scaling} that would require lower-level reimplementation.
That is because self attention itself is broken down to matrix multiplication, an operation that is easily parallelizable on accelerators, and has a myriad of efficient algorithms defined for different use cases in computational software (to name a few: LAPACK, cuBLAS, CUTLASS).
Additionally, most deep learning platforms, such as PyTorch, are written on top of such software, and additional packages (such as cuDNN).
This is very helpful to researchers, as it allows them to use abstractions of operations such as matrix multiplications or convolutions, while the backend decides which algorithm to run based on their hardware, software, and use case. They also typically handle automatic gradient computation, which makes designing and training deep neural networks very straightforward.
\begin{figure}[!t]
    \centering
        \includegraphics[trim={6mm, 6mm, 5mm, 6mm},clip,width=0.475\textwidth]{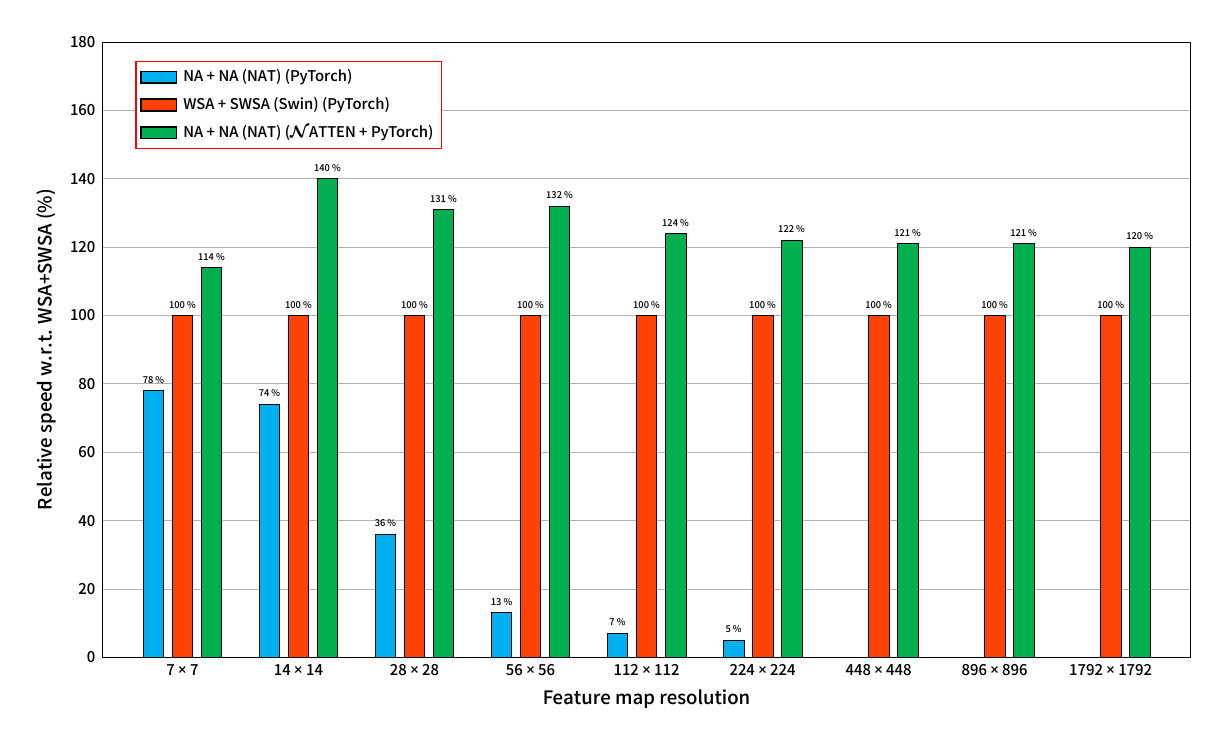}
    \caption{
    \textbf{NAT's layer-wise relative speed with respect to Swin.} 
    Two NA layers with kernel size 7\textsuperscript{2}, are up to 40\% faster than a pair of WSA and SWSA layers with the same kernel size.
    Latency is measured on a single A100 GPU.
    PyTorch implementation of NA runs out of memory at resolutions 448\textsuperscript{2} and higher.
    }
    \label{fig:nattenspeed}
\end{figure}
Because of the pixel-wise structure of NA (and other pixel-wise attention mechanisms, such as SASA~\cite{ramachandran2019stand}), and also the novelty of the definition of neighborhoods in NA, the only way to implement NA with these platforms is to stack a number of highly inefficient operations to extract the neighborhoods, store them as an intermediary tensor, and then compute attention.
This results in a significantly slow operation, with an exponentially growing memory usage.
To tackle these challenges, we developed a set of efficient CPU and CUDA kernels and packaged them as a Python package, \textbf{Neighborhood Attention Extension (\nattenb{})}.
\natten{} includes half precision support, support for both 1D and 2D data, and autograd-compatible integration with PyTorch.
This means that users can simply import NA as a PyTorch module and integrate it into existing pipelines.
We also add that SASA can also be easily implemented with this package with no change in the underlying kernels (simply by padding inputs with zeros), as it is a special case of NA. The reverse does not hold true.
It also includes our \textbf{tiled NA algorithm}, which computes neighborhood attention weights by loading non-overlapping query tiles into shared memory to minimize global memory reads. 
Compared to the naive implementation, tiled NA can decrease latency up to an order of magnitude (see \cref{appdx:natten} for technical details), and it allows NA-based models to run up to 40\% faster than similar Swin counterparts (see \cref{fig:nattenspeed}.)
\natten{} is open-sourced at: \url{https://github.com/SHI-Labs/NATTEN}.

\subsection{Neighborhood Attention Transformer}
\begin{figure*}[!t]
    \centering
    \includegraphics[width=0.96\textwidth]{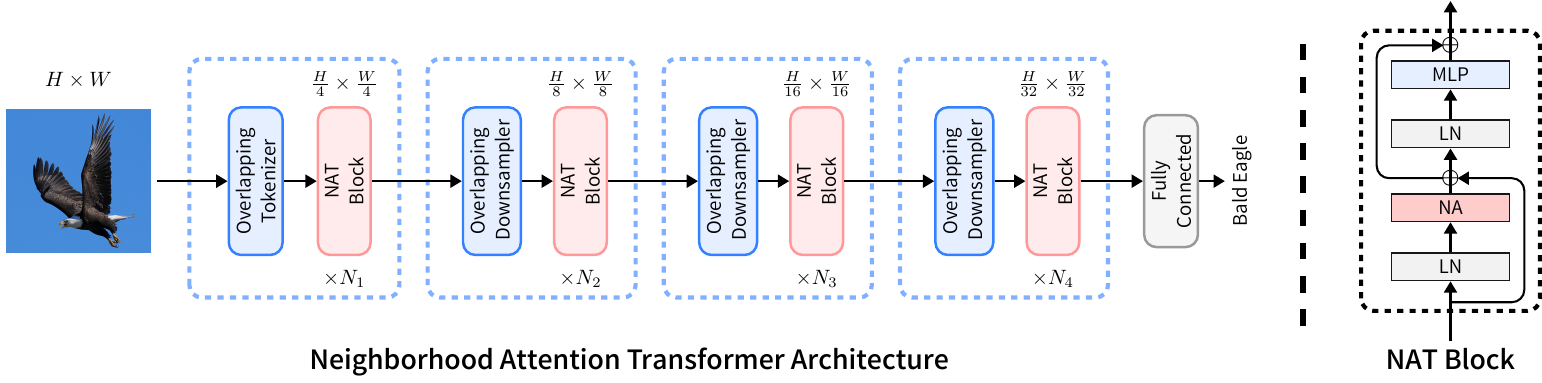}
    \caption{
    \textbf{An overview of our model, NAT, with its hierarchical design.} 
    The model starts off with a convolutional downsampler, then moves on to 4 sequential levels, each consisting of multiple NAT Blocks, which are transformer-like encoder layers. Each layer is comprised of a multi-headed neighborhood attention (NA), a multi-layered perceptron (MLP), Layer Norm (LN) before each module, and skip connections. Between the levels, feature maps are downsampled to half their spatial size, while their depth is doubled. This allows for easier transfer to downstream tasks through feature pyramids. 
    }
    \label{fig:model_overview}
\end{figure*}
NAT embeds inputs using 2 consecutive $3 \times 3$ convolutions with $2 \times 2$ strides, resulting in a spatial size $1/4$th the size of the input. This is similar to using a patch and embedding layer with $4 \times 4$ patches, but it utilizes overlapping convolutions instead of non-overlapping ones to introduce useful inductive biases~\cite{wu2021cvt,hassani2021escaping}. On the other hand, using overlapping convolutions would increase cost, and two convolutions incurs more parameters. However, we handle that by re-configuring the model, which results in a better trade-off.
\setlength{\tabcolsep}{5pt}
\begin{table}[!t]
    \centering
    \resizebox{0.475\textwidth}{!}{
    \begin{tabular}{lccccc}
        \toprule
        \textbf{Variant} & \textbf{Layers} & \textbf{Dim \texttimes{}} & \textbf{MLP} & \textbf{\# of} & \textbf{FLOPs}\\
                         &                 & \textbf{Heads} & \textbf{ratio} & \textbf{Params} &            \\
        \midrule
        \nb\textbf{NAT-Mini}  & 3, 4, \phantom{1}6, 5 & 32 \texttimes{} 2 & 3 & 20 M & \phantom{1}2.7 G \\
        \nb\textbf{NAT-Tiny}  & 3, 4, 18, 5 & 32 \texttimes{} 2 & 3 & 28 M & \phantom{1}4.3 G \\
        \nb\textbf{NAT-Small} & 3, 4, 18, 5 & 32 \texttimes{} 3 & 2 & 51 M & \phantom{1}7.8 G \\
        \nb\textbf{NAT-Base}  & 3, 4, 18, 5 & 32 \texttimes{} 4 & 2 & 90 M & 13.7 G \\
        \bottomrule
    \end{tabular}
    }
    \caption{
    \textbf{Comparison of NAT Variants.}
    }
    \label{tab:nat_variants}
\end{table}
NAT consists of 4 levels, each followed by a downsampler (except the last). Downsamplers cut spatial size in half, while doubling the number of channels. 
We use $3 \times 3$ convolutions with $2 \times 2$ strides, instead of $2 \times 2$ non-overlapping convolutions that Swin uses (patch merge). 
Since the tokenizer downsamples by a factor of $4$, our model produces feature maps of sizes $\frac{h}{4} \times \frac{w}{4}$, $\frac{h}{8} \times \frac{w}{8}$, $\frac{h}{16} \times \frac{w}{16}$, and $\frac{h}{32} \times \frac{w}{32}$.
This change is motivated by previous successful CNN structures, and followed by other hierarchical attention-based methods, such as PVT~\cite{wang2021pyramid}, ViL~\cite{zhang2021multi}, and Swin Transformer~\cite{liu2021swin}.
Additionally, we use LayerScale~\cite{touvron2021going} for stability in larger variants.
An illustration of the overall network architecture is presented in \cref{fig:model_overview}. 
We present a summary of different NAT variants in \cref{tab:nat_variants}.

\subsection{Complexity Analysis}
\label{sec:complexity}
\setlength{\tabcolsep}{3pt}
\begin{table}[!t]
    \centering
    \resizebox{0.475\textwidth}{!}{
    \begin{tabular}{lcc}
        \toprule
        \textbf{Module} & \textbf{FLOPs} & \textbf{Memory}\\
        \midrule
        \vb\textbf{Self Attn (SA)} & $  3 h w d^2 + 2 h^2 w^2 d $ & $ 3 d^2 + h^2 w^2 $ \\
        \wb\textbf{Window Self Attn (WSA)} & $  3 h w d^2 + 2 h w d k^2 $ & $ 3 d^2 + h w k^2 $ \\
        \nb\textbf{Neighborhood Attn (NA)} & $  3 h w d^2 + 2 h w d k^2 $ & $ 3 d^2 + h w k^2 $ \\
        \cb\textbf{Convolution} & $ h w d^2 k^2 $ & $ d^2k^2 $ \\
        \bottomrule
    \end{tabular}
    }
    \caption{
    \textbf{Computational cost and memory usage in different attention patterns and convolutions.}
    SA has a quadratic complexity with respect to resolution, while WSA, NA, and convolutions have a linear complexity.
    }
    \label{tab:complexity_summary}
\end{table}
We present a complexity and memory usage analysis in this subsection, which compares SA, WSA, NA, and convolutions in \cref{tab:complexity_summary}. 
For simplicity, we exclude attention heads. 
Given input feature maps of shape $h \times w \times d$, where $d$ is the number of channels, and $h$ and $w$ are feature map height and width respectively, the $QKV$ linear projections are $3 h w d^2$ FLOPs, which is the same for all three attention patterns.
SA has a quadratic complexity, as both computing attention weights and output are $h^2 w^2 d$ FLOPs, and attention weights are of shape $hw \times hw$.
Swin's WSA divides the queries, keys, and values into $\frac{h}{k} \times \frac{w}{k} $ windows of shape $k \times k$, then applies self attention with each window, which is $ h w d k^2 $ FLOPs. 
WSA's memory consumption, given that its attention weights are of shape $\frac{h}{k} \times \frac{w}{k} \times k^2 \times k^2$, is therefore $ h w d k^2 $.
In NA, $\mathbf{A}_{i}^{k}$ is of size $h \times w \times k^2$, and the cost to compute it is $ h w d k^2 $. 
$\mathbf{V}_{i}^{k}$ is of shape $h \times w \times k^2 \times d$, and therefore the cost of applying attention weights to it would be $ h w d k^2 $.
As for convolutions, computational cost is $ h w d^2 k^2 $, and memory usage would be only $ d^2 k^2 $. 
The summary in \cref{tab:complexity_summary} clarifies that Swin's WSA and NA have identical computational cost and memory usage in theory.
\section{Experiments}
We demonstrate NAT's applicability and effectiveness by conducting experiments across different vision tasks, such as image classification, object detection, and semantic segmentation. 
We also present ablations on different attention patterns, as well as our NAT design.
Additional experiments, including saliency analysis can be found in \cref{appdx:exps}.  

\subsection{Classification}
We trained our variants on ImageNet-1K~\cite{deng2009imagenet} in order to compare to other transformer-based and convolutional image classifiers. This dataset continues to be one of the few benchmarks for medium-scale image classification, containing roughly 1.28M training, 50K validation, and 100K test images, categorized into 1000 classes. 
We train NAT with the commonly used \verb|timm|~\cite{rw2019timm} (Apache License v2), and use the conventional augmentations (CutMix \cite{yun2019cutmix}, Mixup \cite{zhang2017mixup}, RandAugment \cite{cubuk2020randaugment}, and Random Erasing \cite{zhong2020random}) and training techniques used in methods we compare to~\cite{liu2021swin,liu2022convnet}. We follow Swin's~\cite{liu2021swin} training configuration (learning rate, iteration-wise cosine schedule, and other hyperparameters). Following convention, we train for 300 epochs, 20 of which warm up the learning rate, while the rest decay according to the scheduler, and do 10 additional cooldown epochs~\cite{touvron2021training}.
Results are presented in \cref{tab:imagenet_comparison}, and visualized in \cref{fig:natperformance}.
We observe that NAT-Mini outperforms Swin-Tiny by a margin of 0.5\%, with fewer parameters, higher throughput and lower memory usage.
As for the other three variants, we observe they consistently outperform both Swin and ConvNeXt counterparts with similar number of parameters and FLOPs. 
While our Small variant is slightly slower than its Swin counterpart due to the difference in architecture, our Base variant catches up to being faster than Swin-Base.

\setlength{\tabcolsep}{6pt}
\begin{table}[t]
    \centering
    \resizebox{0.475\textwidth}{!}{
        \begin{tabular}{lccccc}
            \toprule
            \textbf{Model} & \textbf{\# of}     & \textbf{FLOPs} & \textbf{Thru.} & \textbf{Memory} & \textbf{Top-1}\\
                           & \textbf{Params}    &                & (imgs/sec)     & (GB)            & (\%) \\
            \midrule
            \ours\textbf{NAT-M}         & 20 M & \phantom{1}2.7 G & 2135 & 2.4 & 81.8 \\
            \midrule
            \wb\textbf{Swin-T}          & 28 M & \phantom{1}4.5 G & 1730 & 4.8 & 81.3 \\
            \cb\textbf{ConvNeXt-T}      & 28 M & \phantom{1}4.5 G & 2491 & 3.4 & 82.1 \\
            \ours\textbf{NAT-T}         & 28 M & \phantom{1}4.3 G & 1541 & 2.5 & \textbf{83.2} \\
            \midrule
            \wb\textbf{Swin-S}          & 50 M & \phantom{1}8.7 G & 1059 & 5.0 & 83.0 \\
            \cb\textbf{ConvNeXt-S}      & 50 M & \phantom{1}8.7 G & 1549 & 3.5 & 83.1 \\
            \ours\textbf{NAT-S}         & 51 M & \phantom{1}7.8 G & 1051 & 3.7 & \textbf{83.7} \\
            \midrule
            \wb\textbf{Swin-B}          & 88 M & 15.4 G & \phantom{1}776 & 6.7 & 83.5 \\
            \cb\textbf{ConvNeXt-B}      & 89 M & 15.4 G & 1107 & 4.8 & 83.8 \\
            \ours\textbf{NAT-B}         & 90 M & 13.7 G & \phantom{1}783 & 5.0 & \textbf{84.3} \\
            \bottomrule
        \end{tabular}
    }
    \caption{
    \textbf{ImageNet-1K classification performance.}
    All models run at 224\texttimes224 resolution with no extra data or pre-training. 
    Peak memory usage and throughput are measured with a batch size of 256 on a single NVIDIA A100 GPU.
    }
    \label{tab:imagenet_comparison}
\end{table}

\subsection{Object Detection and Instance Segmentation}
\label{sec:detection}
We trained Mask~\cite{he2017mask} and Cascade Mask R-CNN~\cite{cai2018cascade} on MS-COCO~\cite{lin2014microsoft}, with NAT backbones, which were pre-trained on ImageNet. We followed Swin~\cite{liu2021swin}'s training settings, using \verb|mmdetection| \cite{chen2019mmdetection} (Apache License v2), and trained with the same accelerated $3\times$ LR schedule. 
Results are presented in \cref{tab:cocodetectioninstance}. 
NAT-Mini outperforms Swin-Tiny with Mask R-CNN, while falling slightly short to it with Cascade Mask R-CNN, all while having significantly fewer FLOPs. 
NAT-Tiny outperforms both its Swin and ConvNeXt counterparts, with both Mask and Cascade Mask R-CNN, while having a slightly lower throughput compared to its Swin counterpart. 
NAT-Small reaches a competitive performance compared to its Swin counterpart, while being faster.
NAT-Base can even outperform its Swin counterpart, while also enjoying a higher throughput.

\setlength{\tabcolsep}{3pt}
\begin{table}[t]
    \centering
    \resizebox{0.475\textwidth}{!}{
        \begin{tabular}{lccc|ccc|ccc}
            \toprule
            \textbf{Backbone} & \textbf{\# of} & \textbf{FLOPs} & \textbf{Thru.} & \textbf{AP\textsuperscript{b}} & \textbf{AP\textsuperscript{b}\textsubscript{50}} & \textbf{AP\textsuperscript{b}\textsubscript{75}} & \textbf{AP\textsuperscript{m}} & \textbf{AP\textsuperscript{m}\textsubscript{50}} & \textbf{AP\textsuperscript{m}\textsubscript{75}} \\ 
            & \textbf{Params} &&(FPS)\\
            \midrule
            \multicolumn{10}{c}{\textit{Mask R-CNN - 3x schedule}} \\
            \midrule
            \ours\textbf{NAT-M}         &\phantom{1}40 M & 225 G & 54.1 & 46.5 & 68.1 & 51.3 & 41.7 & 65.2 & 44.7 \\
            \midrule
            \wb\textbf{Swin-T}          &\phantom{1}48 M & 267 G & 45.1 & 46.0 & 68.1 & 50.3 & 41.6 & 65.1 & 44.9 \\
            \cb\textbf{ConvNeXt-T}      &\phantom{1}48 M & 262 G & 52.0 & 46.2 & 67.0 & 50.8 & 41.7 & 65.0 & 44.9 \\
            \ours\textbf{NAT-T}         &\phantom{1}48 M & 258 G & 44.5 & 47.7 & 69.0 & 52.6 & 42.6 & 66.1 & 45.9 \\
            \midrule
            \wb\textbf{Swin-S}          &\phantom{1}69 M & 359 G & 31.7 & 48.5 & 70.2 & 53.5 & 43.3 & 67.3 & 46.6 \\
            \ours\textbf{NAT-S}         &\phantom{1}70 M & 330 G & 34.8 & 48.4 & 69.8 & 53.2 & 43.2 & 66.9 & 46.5 \\
            \midrule
            \multicolumn{10}{c}{\textit{Cascade Mask R-CNN - 3x schedule}} \\
            \midrule
            \ours\textbf{NAT-M}         & \phantom{1}77 M & 704 G & 27.8 & 50.3 & 68.9 & 54.9 & 43.6 & 66.4 & 47.2 \\
            \midrule
            \wb\textbf{Swin-T}          & \phantom{1}86 M & 745 G & 25.1 & 50.4 & 69.2 & 54.7 & 43.7 & 66.6 & 47.3 \\
            \cb\textbf{ConvNeXt-T}      & \phantom{1}86 M & 741 G & 27.3 & 50.4 & 69.1 & 54.8 & 43.7 & 66.5 & 47.3 \\
            \ours\textbf{NAT-T}         & \phantom{1}85 M & 737 G & 24.9 & 51.4 & 70.0 & 55.9 & 44.5 & 67.6 & 47.9 \\
            \midrule
            \wb\textbf{Swin-S}          & 107 M & 838 G & 20.3 & 51.9 & 70.7 & 56.3 & 45.0 & 68.2 & 48.8 \\
            \cb\textbf{ConvNeXt-S}      & 108 M & 827 G & 23.0 & 51.9 & 70.8 & 56.5 & 45.0 & 68.4 & 49.1 \\
            \ours\textbf{NAT-S}         & 108 M & 809 G & 21.7 & 52.0 & 70.4 & 56.3 & 44.9 & 68.1 & 48.6 \\
            \midrule
            \wb\textbf{Swin-B}          & 145 M & 982 G & 17.3 & 51.9 & 70.5 & 56.4 & 45.0 & 68.1 & 48.9 \\
            \cb\textbf{ConvNeXt-B}      & 146 M & 964 G & 19.5 & 52.7 & 71.3 & 57.2 & 45.6 & 68.9 & 49.5 \\
            \ours\textbf{NAT-B}         & 147 M & 931 G & 18.6 & 52.5 & 71.1 & 57.1 & 45.2 & 68.6 & 49.0 \\
            \bottomrule
        \end{tabular}
    }
    \caption{
    \textbf{COCO object detection and instance segmentation performance.} 
    FLOPS are with respect to an input resolution of (1280, 800). 
    Throughput is measured at the same resolution on a single NVIDIA A100 GPU.
    }
    \label{tab:cocodetectioninstance}
\end{table}

\subsection{Semantic Segmentation}
\label{sec:segmentation}

\setlength{\tabcolsep}{8pt}
\begin{table}[t]
    \centering
    \resizebox{0.475\textwidth}{!}{
        \begin{tabular}{lccc|cc}
            \toprule
            \textbf{Backbone} & \textbf{\# of}      & \textbf{FLOPs} & \textbf{Thru.} & \multicolumn{2}{c}{\textbf{mIoU}} \\ 
                              & \textbf{Params}     &                & (FPS)          & single scale & multi scale        \\
            \midrule
            \ours\textbf{NAT-M}         & \phantom{1}50 M & \phantom{1}900 G & 24.5 & 45.1 & 46.4 \\
            \midrule
            \wb\textbf{Swin-T}          & \phantom{1}60 M & \phantom{1}946 G & 21.3 & 44.5 & 45.8 \\
            \cb\textbf{ConvNeXt-T}      & \phantom{1}60 M & \phantom{1}939 G & 23.3 & 46.0 & 46.7 \\
            \ours\textbf{NAT-T}         & \phantom{1}58 M & \phantom{1}934 G & 21.4 & 47.1 & 48.4 \\
            \midrule
            \wb\textbf{Swin-S}          & \phantom{1}81 M & 1040 G & 17.0 & 47.6 & 49.5 \\
            \cb\textbf{ConvNeXt-S}      & \phantom{1}82 M & 1027 G & 19.1 & 48.7 & 49.6 \\
            \ours\textbf{NAT-S}         & \phantom{1}82 M & 1010 G & 17.9 & 48.0 & 49.5 \\
            \midrule
            \wb\textbf{Swin-B}          & 121 M & 1188 G & 14.6 & 48.1 & 49.7 \\
            \cb\textbf{ConvNeXt-B}      & 122 M & 1170 G & 16.4 & 49.1 & 49.9 \\
            \ours\textbf{NAT-B}         & 123 M & 1137 G & 15.6 & 48.5 & 49.7 \\
            \bottomrule
        \end{tabular}
    }
    \caption{
    \textbf{ADE20K semantic segmentation performance.} 
    FLOPS are with respect to an input resolution of (2048, 512). 
    Throughput is measured at the same resolution on a single NVIDIA A100 GPU.
    }
    \label{tab:semseg}
\end{table}
To demonstrate NAT's performance on semantic segmentation, we trained UPerNet~\cite{xiao2018unified} with NAT backbones on ADE20K~\cite{zhou2017scene}. 
We followed Swin's configuration for training ADE20K, and used \verb|mmsegmentation|~\cite{mmseg2020} (Apache License v2). 
Additionally, and following standard practice, input images are randomly resized and cropped at $512 \times 512$ when training. 
Results are presented in \cref{tab:semseg}.
It is noticeable that NAT-Mini outperforms Swin-Tiny, and also comes very close to ConvNeXt-Tiny. NAT-Tiny outperforms ConvNeXt-Tiny significantly, while also slightly more efficient. NAT-Small outperforms Swin-Small on single-scale performance, while matching the multi-scale performance. NAT-Base similarly performs on-par with Swin-Base, while falling slightly short of ConvNeXt-Base. 
Note that both NAT-Small and NAT-Base bear fewer FLOPs with them compared to their Swin and ConvNeXt counterparts, while their performance is within the same region.
It is also noteworthy that Swin especially suffers from more FLOPs even beyond the original difference due to the fact that the image resolution input in this task specifically ($512 \times 512$) will not result in feature maps that are divisible by $7 \times 7$, Swin's window size, which forces the model to pad input feature maps with zeros to resolve that issue, prior to every attention operation. 
NAT on the other hand supports feature maps of arbitrary size.

\subsection{Ablation Study}
We compare Swin's attention pattern (WSA+SASA) to sliding window patterns, namely SASA~\cite{ramachandran2019stand} (implemented with our \natten{} package and therefore enjoys approximately the same throughput and identical memory usage as NA), and our NA.
We simply replace the attention blocks in Swin-Tiny, and run the model on ImageNet-1K classification, MSCOCO object detection and instance segmentation, and ADE20K segmentation.
Results are presented in \cref{tab:attentionablation}.

Separately, we investigate the effects of our NAT design (convolutional downsampling and deeper-thinner architecture), by performing an ablation with Swin-Tiny as baseline. We slowly transform the model into NAT-Tiny, and present the results in \cref{tab:modelablation}.
We start by replacing the patched embedding and patched merge with the overlapping convolution design used in NAT.
This results in almost 0.5\% improvement in accuracy. After taking the second step to reduce the model size and compute, by making it deeper and thinner, we notice the model sees approximately an improvement in accuracy of 0.9\% over the first step.
We then try swapping the WSA and SWSA attention patterns in Swin with SASA~\cite{ramachandran2019stand}, and see a slight drop in accuracy.
However, swapping WSA and SWSA with our NA shows a further 0.5\% improvement in accuracy.

We also present a kernel size experiment in \cref{tab:kernelsizecomparison}, in which we try kernel sizes ranging from 3\texttimes{} 3 to 9 \texttimes{} 9, in an effort to analyze its affect on our model's performance.

\setlength{\tabcolsep}{1pt}
\begin{table}[t]
    \centering
    \resizebox{0.475\textwidth}{!}{
    \begin{tabular}{l|c|cc|c|cccc}
        \toprule
        \textbf{Attention}                      & \textbf{ImageNet} & \multicolumn{2}{c|}{\textbf{MSCOCO}} & \textbf{ADE20K} & \textbf{\# of} & \textbf{FLOPs} & \textbf{Thru.} & \textbf{Memory}\\
                                                & \textbf{Top-1} & \textbf{AP\textsuperscript{B}} & \textbf{AP\textsuperscript{m}} & \textbf{mIoU} & \textbf{Params} &  & (imgs/sec) & (GB)\\
        \midrule
        \wb\textbf{SWSA}    & 81.3\% & 46.0 & 41.6 & 45.8 & 28.28 M & 4.51 G & 1730 & 4.8 \\
        \sab\textbf{SASA}  & 81.6\% & 46.0 & 41.4 & 46.4 & 28.27 M & 4.51 G & 2021 & 4.0 \\
        \ours\textbf{NA}     & 81.8\% & 46.2 & 41.5 & 46.4 & 28.28 M & 4.51 G & 2021 & 4.0 \\
        \bottomrule
    \end{tabular}}
    \caption{
    \textbf{Performance comparison of different attention mechanisms.} 
    All models are presented are identical in architecture to Swin-T, with only the attention mechanisms replaced.
    SASA is implemented with our \natten{}, and therefore enjoys the same speed and memory efficiency as NA.
    }
    \label{tab:attentionablation}
\end{table}

\setlength{\tabcolsep}{1.5pt}
\begin{table}[!ht]
    \centering
    \resizebox{0.475\textwidth}{!}{
        \begin{tabular}{lccccc|cccc}
            \toprule
            \textbf{Attention} & \textbf{Down-} & \textbf{\# of}     & \textbf{\# of} & \textbf{MLP}     & \textbf{Top-1} & \textbf{\# of}   & \textbf{FLOPs} & \textbf{Thru.} & \textbf{Memory}\\
                               & \textbf{-sampler} & \textbf{Layers}    & \textbf{Heads} & \textbf{Ratio}   & (\%)           & \textbf{Params}  & (G) & (imgs/sec) & (GB)\\
            \midrule
            \wb\textbf{SWSA}    & Patch &  2, 2,\phantom{1}6, 2 & 3 & 4 & 81.29 & 28.3 M & 4.5 & 1730 & 4.8\\
            \midrule
            \wb\textbf{SWSA}    & Conv  &  2, 2,\phantom{1}6, 2 & 3 & 4 & 81.78 & 30.3 M & 4.9 & 1692 & 4.8 \\
            \nb\textbf{SWSA}    & Conv  & 3, 4, 18, 5  & 2 & 3 & 82.72 & 27.9 M & 4.3 & 1320 & 3.0 \\
            \midrule
            \sab\textbf{SASA}   & Conv  & 3, 4, 18, 5 & 2 & 3 & 82.54 & 27.9 M & 4.3 & 1541 & 2.5 \\
            \midrule
            \ours\textbf{NA}    & Conv  & 3, 4, 18, 5 & 2 & 3 & 83.20 & 27.9 M & 4.3 & 1541 & 2.5 \\
            \bottomrule
        \end{tabular}
    }
    \caption{
    \textbf{Ablation study on NAT, with Swin-T as the baseline.}
    Through overlapping convolutions (row 2), and our NAT configuration (row 3), we boost Swin classification accuracy (row 1) significantly. 
    Swapping SWSA with NA (row 5) results in an improvement of almost 0.5\% in accuracy, while swapping it with SASA (row 4) results in a slight decrease in accuracy.
    SASA is implemented with our \natten{}, and therefore enjoys the same speed and memory efficiency as NA.
    }
    \label{tab:modelablation}
\end{table}

\setlength{\tabcolsep}{3pt}
\begin{table}[t]
    \centering
    \resizebox{0.475\textwidth}{!}{
    \begin{tabular}{l|cc|ccc|cc}
        \toprule
        \textbf{Kernel}         & \multicolumn{2}{c|}{\textbf{ImageNet}} & \multicolumn{3}{c|}{\textbf{MSCOCO}}                                               & \multicolumn{2}{c}{\textbf{ADE20K}}  \\
        \textbf{size}           & \textbf{Top-1 (\%)}& \textbf{Thru.}     & \textbf{AP\textsuperscript{b}} & \textbf{AP\textsuperscript{m}}   & \textbf{Thru.} & \textbf{mIoU} & \textbf{Thru.}       \\
        \midrule
        3\texttimes{}3          & 81.4   & 2015 \ips         & 46.1                           & 41.4                             & 46.8 \fps      & 46.0          & 23.6 \fps            \\
        5\texttimes{}5          & 81.6   & 1810 \ips         & 46.8                           & 42.0                             & 45.5 \fps      & 46.3          & 22.9 \fps            \\
        \grayrow 7\texttimes{}7 & 83.2   & 1537 \ips         & 47.7                           & 42.6                             & 44.5 \fps      & 48.4          & 21.4 \fps            \\
        9\texttimes{}9          & 83.1   & 1253 \ips         & 48.5                           & 43.3                             & 39.4 \fps      & 48.1          & 20.2 \fps            \\
        \bottomrule
    \end{tabular}
    }
    \caption{
    \textbf{NAT-Tiny performance with different kernel sizes.}
    }
    \label{tab:kernelsizecomparison}
\end{table}
\section{Conclusion}
\label{sec:conclusion}
In this paper, we present Neighborhood Attention (NA), the first efficient and scalable sliding window attention mechanism for vision.
NA is a pixel-wise operation which localizes self attention for each pixel to its nearest neighborhood, and therefore enjoys linear complexity. 
It also introduces local inductive biases and maintains translational equivariance, unlike blocked (HaloNet) and window self attention (Swin).
Different from SASA, NA approaches self attention as its window size grows, and does not limit attention span at corner cases.
We challenge the common notion that explicit sliding window attention patterns are not efficient or parallelizable~\cite{liu2021swin} by developing \natten{}.
Through using \natten{}, NA-based models can run even faster than existing alternatives, despite the latter running primarily on highly optimized deep learning libraries built on top of lower-level computational packages.
We also propose Neighborhood Attention Transformer (NAT) and show the power of such models: NAT outperforms both Swin Transformer and ConvNeXt in image classification, and outperforms or competes with both in downstream vision tasks. 
We open-source our entire project to encourage more research in this direction.

\paragraph{Acknowledgments.} 
This work was in part supported by the Intelligence Advanced Research Projects Activity (IARPA) under Contract No. 2022-21102100004.
We thank Picsart AI Research (PAIR) and Meta for their generous support that made this work possible.

{\small
\bibliographystyle{ieee_fullname}
\bibliography{references}
}

\clearpage 
\appendix
\renewcommand{\thetable}{\Roman{table}}
\renewcommand{\thefigure}{\Roman{figure}}
\setcounter{table}{0}
\setcounter{figure}{0}

\section*{\Large{Appendix}}
\begin{figure*}[!ht]
    \centering
    \includegraphics[width=0.99\textwidth]{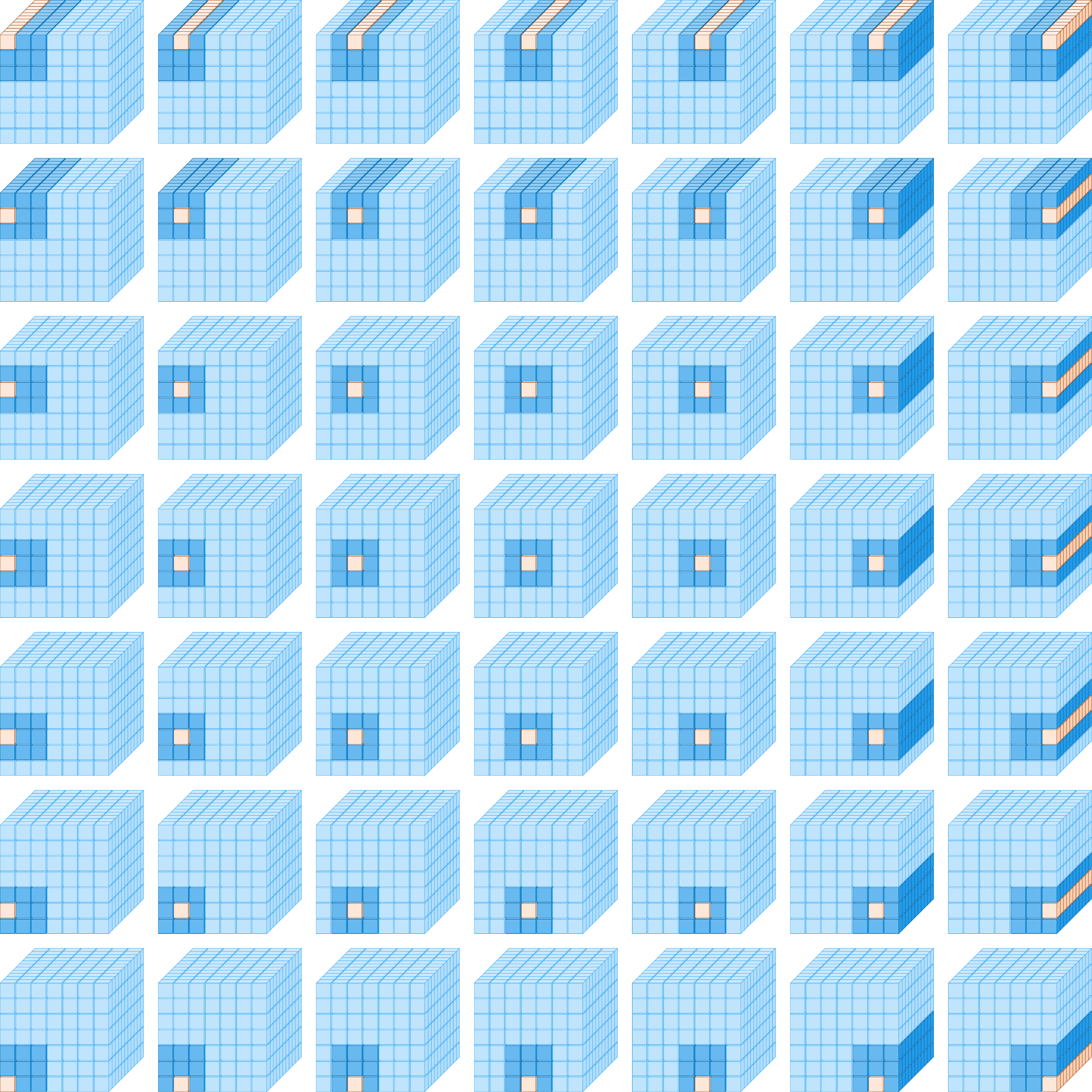}
    \caption{
    \textbf{An illustration of 3 \texttimes{} 3 neighborhood attention pattern on a 7 \texttimes{} 7 feature map.}
    Query is colored orange, and its attention span (key-value pair) is dark blue.
    The ``window'' is repeated at the corners because of the neighborhood definition. 
    This keeps attention span identical to the rest of the feature map. 
    The alternative to this would have been smaller neighborhoods (zero padding at the corners, similar to SASA). 
    }
    \label{appfig:fullneighborhood}
\end{figure*}

We present a more detailed illustration of neighborhoods in \cref{appfig:fullneighborhood}.
Note that the repeated windows at corner pixels is especially important to NA approaching SA.
We also present details on our \natten{} package in \cref{appdx:natten}, and additional experiments in \cref{appdx:exps}.
Additionally, we discuss translational equivariance in self attention mechanisms in \cref{appdx:te}.

\section{\texorpdfstring{\nattenb}{NATTEN}}
\label{appdx:natten}

In this section, we outline the necessity for an extension such as \natten{} for research in the direction of dynamic sliding window attention patterns, and describe how it aims to resolve such problems.

\subsection{Background}
\begin{figure*}[!t]
    \centering
    \includegraphics[trim={6mm, 6mm, 5mm, 5mm},clip,width=\textwidth]{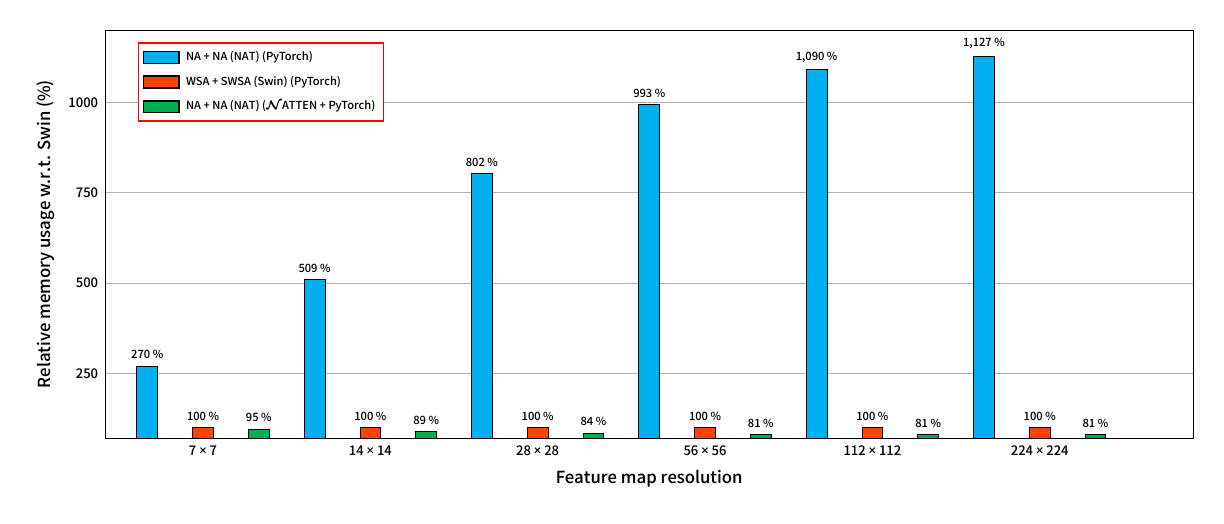}
    \caption{
    \textbf{NAT's layer-wise memory usage with respect to Swin.} 
    Without \natten{}, a plain PyTorch implementation of NA bears a quickly growing memory footprint compared to Swin, whereas with \natten{}, it uses consistently lower memory.
    }
    \label{appfig:pymemory}
\end{figure*}

While many operations in deep neural networks can be broken down to matrix multiplication, certain point-wise operations, such as convolutions, require customized implementations for more optimal parallelization.
As a result, convolutions, recurrent modules, and other similar operations are natively supported in most low-level computational packages, which are called by deep learning frameworks such as PyTorch.
In other words, given any input set and an operation, deep learning frameworks select the most efficient implementation available for that particular case, considering the hardware and software running said operation.

This makes research significantly easier, while being inevitably constrained to operations that are well-implemented.
To allow further flexibility, some deep learning frameworks also allow for extensions to be built on top of them when necessary.
Extensions can therefore enjoy customized CPU and GPU implementations.
Notable examples of such extensions are Deformable Convolutions~\cite{dai2017deformable} and Deformable Attention~\cite{zhu2020deformable}, which have been implemented as CUDA extensions to PyTorch.

\begin{figure*}[!t]
    \centering
    \includegraphics[trim={6mm, 6mm, 5mm, 6mm},clip,width=\textwidth]{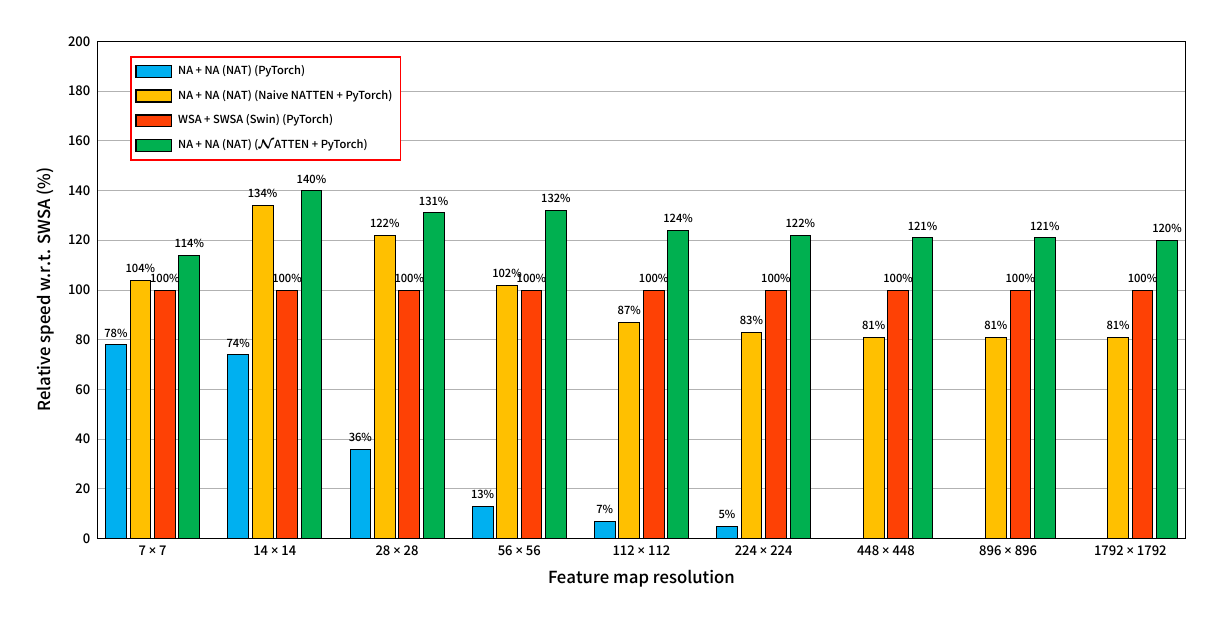}
    \caption{
    \textbf{Torch-based NA, Naive NA, and Tiled NA relative throughput comparison w.r.t. WSA+SWSA.} 
    Latency is measured on a single A100 GPU.
    Note that the plain PyTorch implementation of NA runs out of memory for resolutions 448\textsuperscript{2} and higher.
    }
    \label{appfig:cudatime}
\end{figure*}

Sliding window attention mechanisms are no different, in that they require manual implementation to maximize parallelization and bandwidth.
Without those implementations, the only alternative is a Python implementation, which typically does not scale.
For instance, implementing Neighborhood Attention with PyTorch alone would include extracting sliding windows, repeating and re-arranging them to produce neighborhoods, and then performing two batched matrix multiplications.
This would mean two separate C++/CUDA calls to generate significantly large intermediary tensors, which result in an exponential memory usage and latency increases.
With the most optimized plain PyTorch implementation, NA would run at \textbf{13\% the speed of Swin Transformer} on a 56 \texttimes{} 56 feature map (first level of an ImageNet model), while using approximately \textbf{9 times as much memory}. 
With a naive CUDA implementation, the same NA module runs at \textbf{102\% the speed of Swin}, while using approximately \textbf{20\% less memory}.
With our \textbf{Tiled NA} algorithm, that same module runs at \textbf{132\% the speed of Swin}, with no change in memory usage.
You can refer to \cref{appfig:cudatime,appfig:pymemory} for more benchmarks comparing different NA implementations to Swin Transformer in terms of relative speed and memory usage.

This is why we developed \natten{}, which currently serves as an extension to PyTorch, and provides torch modules \verb|NeighborhoodAttention1D| and \verb|NeighborhoodAttention2D|.
This allows any PyTorch user to integrate NA into their models, for both tokens and pixels.

Each module consists of linear projections for queries, keys, and values, and a final linear projection, which is standard in most dot-product self attention modules.
\natten{} provides a single autograd function for each of \cref{eq:nattenq} and \cref{eq:natten}.
Once tensors \verb|q|, \verb|k|, and \verb|v| are generated, attention weights are computed (see \cref{eq:nattenq}) by passing \verb|q|, \verb|k|, and positional biases \verb|b| to the C function \verb|QK+RPB|, which picks the appropriate kernel to call (CPU or CUDA; naive or special; half or full precision). 
Softmax and dropout are then applied to the output attention weights, \verb|a|, with native torch implementations.
NA's final is computed by passing \verb|a| and \verb|v| to the C function \verb|AV|.

\begin{figure*}[t]
    \centering
    \includegraphics[trim={6mm, 6mm, 5mm, 5mm},clip,width=\textwidth]{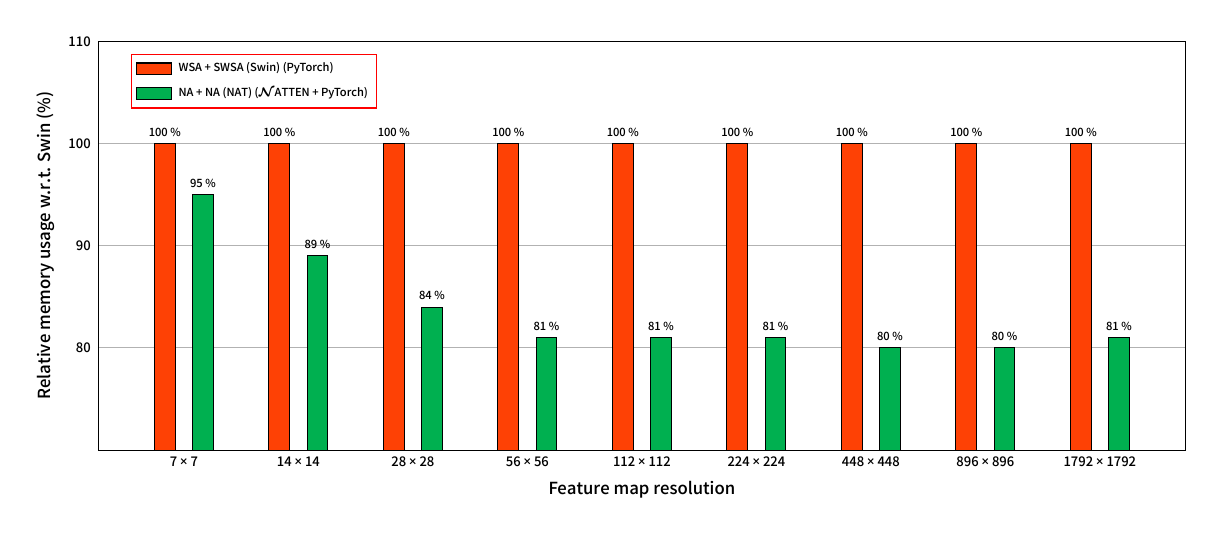}
    \caption{
    \textbf{NAT's layer-wise memory usage with respect to Swin.} 
    Since NA does not include a pixel shift and masked attention like SWSA, and the addition of positional biases is fused into the C++/CUDA kernels, NA with \natten{} uses less memory compared to a similar model with WSA+SWSA.
    }
    \label{appfig:cudamemory}
\end{figure*}

\subsection{Naive CUDA Kernels}
Originally, we developed 7 naive CUDA kernels: 1 for \verb|QK+RPB|, 1 for \verb|AV|, and 5 to compute gradients for each of \verb|q|, \verb|k|, \verb|b|, \verb|a|, and \verb|v|.
Naive kernels simply divide computation across available threadblocks, and do not utilize shared memory or warp optimization.
Despite their simplicity, they were able to benchmark between 80\% up to 130\% the speed of WSA+SWSA layers (both with kernel size 7 \texttimes{} 7).
However, naive kernels are not optimal; they read directly from the global memory on the GPU, which bottlenecks throughput.

\subsection{Half precision}
Supporting mixed precision training is not too complicated. PyTorch's ATen dispatchers compile all kernels for both double and single precision by default, since tensor data type is usually templated.
By choosing a different dispatcher, kernels can be easily compiled for \verb|half| tensors.
However, simply support half precision rarely results in any significant bandwidth improvement without integrating CUDA's vectorized \verb|half2| data type and operators.
As a result, we separately define our half precision kernels to utilize vectorize load, multiply-add, and stores.
This yields a more significant improvement in mixed-precision training speed.

\subsection{Tiled Neighborhood Attention}

\begin{figure}[!t]
    \centering
    \includegraphics[width=0.475\textwidth]{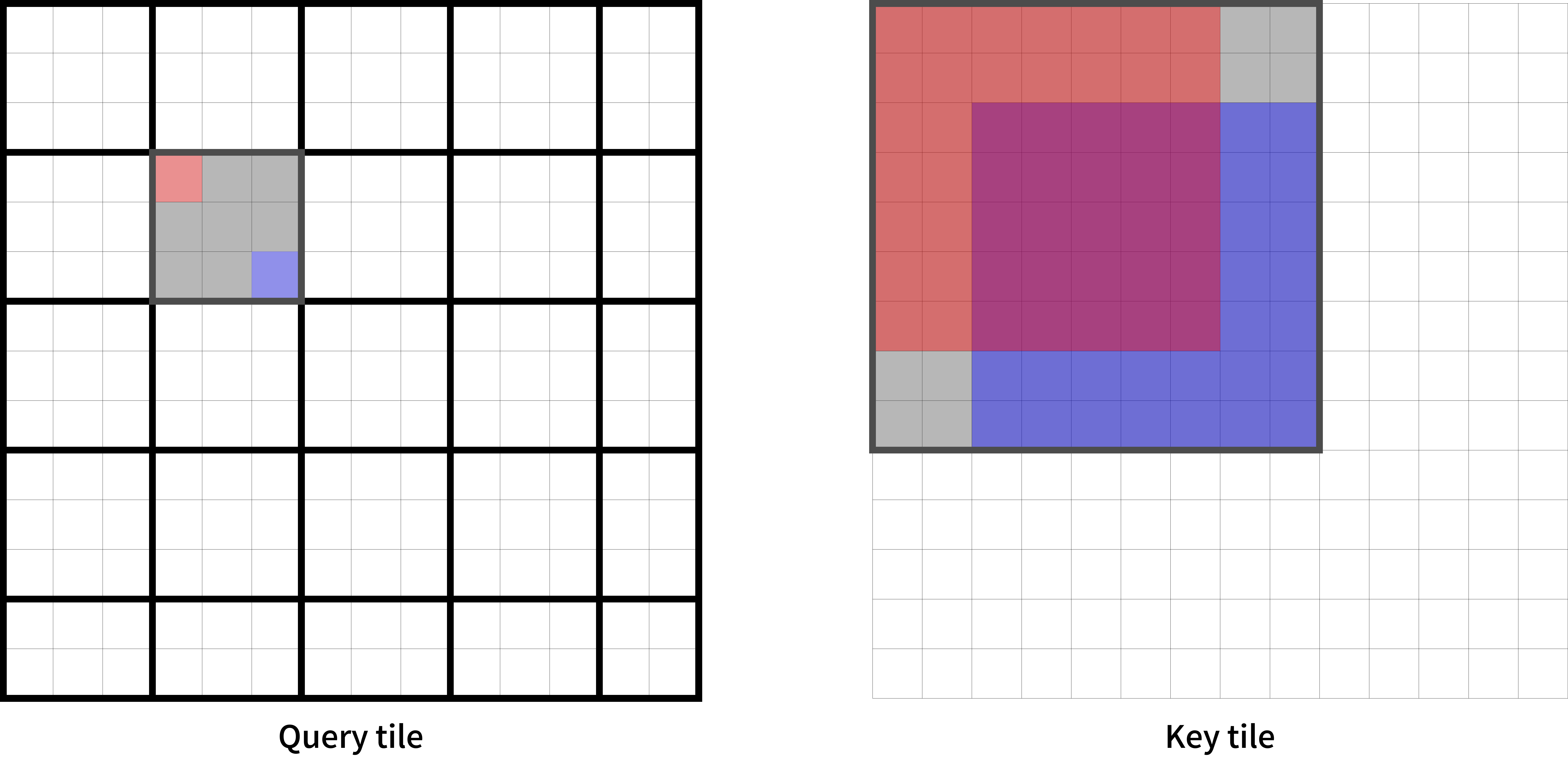}
    \caption{
    \textbf{An illustration of tiled neighborhood attention for kernel size 7 \texttimes{} 7 and tile size 3 \texttimes{} 3. } 
    Queries are partitioned into tiles (left), and because of the large overlap in neighboring keys, it is easy to predict a tile in keys based on kernel size (right).
    This allows each separate threadblock, which has a shared memory between threads, to compute outputs for a specific query tile.
    Two queries (top left and bottom right) and their respective neighborhoods are also highlighted with different colors to visualize that the information needed to compute outputs for each tile is available in the tiles that are loaded.
    }
    \label{fig:tiledna}
\end{figure}

CUDA allows easy allocation and utilization of shared memory between threadblocks. 
This, however, typically requires a change in the algorithm.
Therefore, we implemented a tiled version of our attention weight kernel, and its backward kernel, which divides inputs into non-overlapping tiles, assigns each thread within the threadblock to read a specific number of adjacent cells from global memory, sync, and then compute outputs based on values in the shared memory.
We present an illustration of that in \cref{fig:tiledna}.
Using shared memory also presents new challenges, including, but not limited to: 
1. Tile size bounds depending on kernel size, dimension, and shared memory limit on the GPU. 
2. Bank conflicts between warps during computation. 
3. Different number of reads from each input depending on tile size.

For instance, \cref{fig:tiledna} illustrates NA at kernel size 7 \texttimes 7, with tile size 3 \texttimes{} 3, which requires a key tile of size 9 \texttimes{} 9.
The 3 \texttimes {} 3 tile size was chosen based on a number of factors, including the size of shared memory (48 KB), total number of threads per threadblock (1024 since compute capability 2.0), and other problem-specific factors such as embedding dimension.
Key tile size is always equal to $t_q + k - 1$, where $t_q$ is the query tile size, and $k$ is kernel size, which is $3 + 7 - 1 = 9$ here.

Through a detailed internal analysis, we implemented and optimized Tiled NA for kernel sizes 3, 5, 7, 9, 11, and 13.
Although not all bank conflicts were avoided in all use cases, they were minimized through profiling with NVIDIA Nsight\textsuperscript{TM}.
Even though this implementation has resulted in a considerable bandwidth increase in NA training and inference, \natten{} is still fairly at an early stage.
We hope to improve existing kernels and add more optimal ones for different use cases, and add support for the new Hopper architecture with CUDA 12.

\subsection{CPU Kernels}
We extend  \natten{} to support CPU operations as well, both for training and inference.
CPU functions for Neighborhood Attention are simple C++ implementations, with AVX vectorization support in newer PyTorch versions.
As a result, they can easily utilize multi-threaded computation, which usually results in a relatively good latency compared to similar sized models on consumer CPUs.
In total, there are 7 CPU kernels in the current version (similar to the naive implementations, 1 for each operation and 1 for each gradient.)
We foresee further optimizations and additional CPU kernels in the near future.

\subsection{Future efforts}
We hope to continue supporting \natten{} and help the community enjoy sliding window attention modules. Our hope is to eventually implement Neighborhood Attention with implicit GEMM (generalized matrix-matrix product), which will allow \natten{} to be built on top of open-source packages (i.e. CUTLASS) and utilize the power of hardware accelerators to a greater extent.

\section{Additional experiments}
\label{appdx:exps}

\subsection{Ablation on RPB}
We present an ablation on relative positional biases and pixel shifts (WSA only) in \cref{appdxtab:rpb}.

\setlength{\tabcolsep}{3pt}
\begin{table}[!ht]
    \centering
    \resizebox{0.475\textwidth}{!}{
        \begin{tabular}{lcccc}
            \toprule
            \textbf{Attention}          & \textbf{Positional} & \textbf{Top-1} & \textbf{\# of} & \textbf{FLOPs}\\
                                        & \textbf{biases}            & (\%)           & \textbf{Params}\\
            \midrule
            \wb\textbf{WSA-SWSA}        & None                  & 80.1 (+ 0.0) & 28.26 M & 4.51 G \\
            \ours\textbf{NA+NA}                                     & None                  & 80.6 (+ 0.5) & 28.26 M & 4.51 G \\
            \midrule
            \wb\textbf{WSA-WSA}         & Relative Pos. Bias.   & 80.2 (+ 0.0) & 28.28 M & 4.51 G \\
            \wb\textbf{WSA-SWSA}        & Relative Pos. Bias.   & 81.3 (+ 1.1) & 28.28 M & 4.51 G \\
            \sab\textbf{SASA-SASA}      & Relative Pos. Bias.   & 81.6 (+ 0.3) & 28.28 M & 4.51 G \\
            \ours\textbf{NA-NA}         & Relative Pos. Bias.   & 81.8 (+ 0.5) & 28.28 M & 4.51 G \\
            \bottomrule
        \end{tabular}
    }
    \caption{
    \textbf{Comparing NA and WSA with and without positional biases.}
    Swin's results are directly reported from the original paper.
    }
    \label{appdxtab:rpb}
\end{table}

\subsection{Saliency analysis}
\begin{figure*}
    \begin{center}
        \newcommand{\smimgheight}{3.225cm} 
        \setlength{\fboxsep}{0pt} 
        \setlength\tabcolsep{1pt} 
        \renewcommand{\arraystretch}{0.6667} 
            \begin{tabular}{cccc}
                Original & ViT-B & Swin-B & NAT-B  \\
                \fbox{\includegraphics[height=\smimgheight]{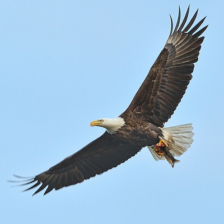}} & \fbox{\includegraphics[height=\smimgheight]{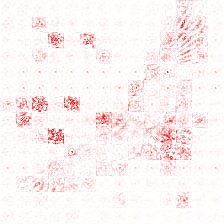}} & \fbox{\includegraphics[height=\smimgheight]{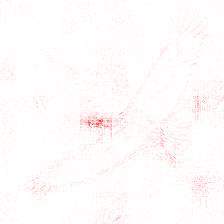}} & \fbox{\includegraphics[height=\smimgheight]{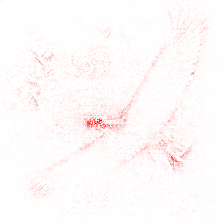}} \\
                \fbox{\includegraphics[height=\smimgheight]{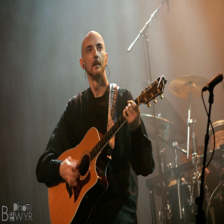}} & \fbox{\includegraphics[height=\smimgheight]{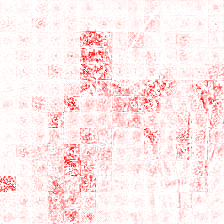}} & \fbox{\includegraphics[height=\smimgheight]{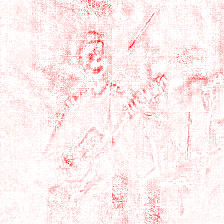}} & \fbox{\includegraphics[height=\smimgheight]{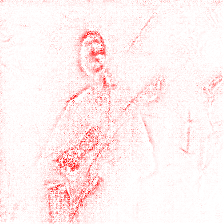}} \\
                \fbox{\includegraphics[height=\smimgheight]{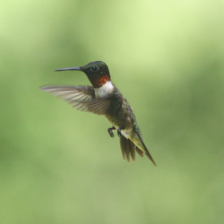}} & \fbox{\includegraphics[height=\smimgheight]{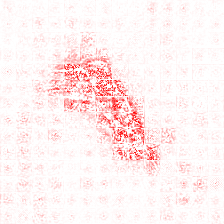}} & \fbox{\includegraphics[height=\smimgheight]{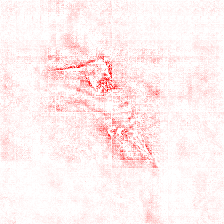}} & \fbox{\includegraphics[height=\smimgheight]{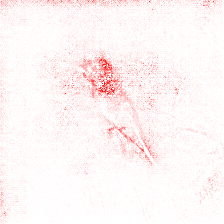}} \\
                \fbox{\includegraphics[height=\smimgheight]{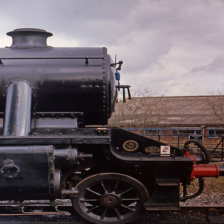}} & \fbox{\includegraphics[height=\smimgheight]{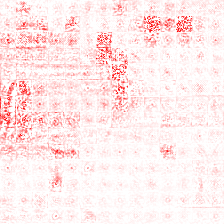}} & \fbox{\includegraphics[height=\smimgheight]{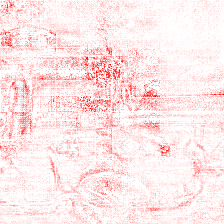}} & \fbox{\includegraphics[height=\smimgheight]{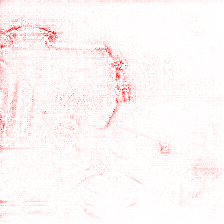}} \\
            \end{tabular}
            \caption{
            \textbf{Salient maps of selected ImageNet validation set images, comparing ViT-Base, Swin-Base, and NAT-Base.} 
            The ground truths for these images are: Bald Eagle, Acoustic Guitar, Hummingbird, and Steam Locomotive, respectively. 
            }
            \label{appfig:salient}
            \setlength\tabcolsep{6pt} 
        \renewcommand{\arraystretch}{1} 
        \setlength{\fboxsep}{3pt} 
    \end{center}
\end{figure*}

\looseness=-1 In an effort to further illustrate the differences between attention mechanisms and models, we present salient maps from ViT-Base, Swin-Base, and NAT-Base. We selected a few images from the ImageNet validation set, sent them through the three models, and created the salient maps based on the outputs, which are presented in \cref{appfig:salient}. All images are correctly predicted (Bald Eagle, Acoustic Guitar, Hummingbird, Steam Locomotive) except ViT's Acoustic Guitar which predicts Stage. From these salient maps we can see that all models have relatively good interpretability, though they focus on slightly different areas. NAT appears to be slightly better at edge detection, which we believe is due to the localized attention mechanism, that we have presented in this work, as well as the convolutional downsamplers.

\section{Notes on translational equivariance}
\label{appdx:te}
In this section, we discuss the translational equivariance property in attention-based models, which is often referenced as a useful property in convolutional models~\cite{goodfellow2016deep}.
To do that, we begin with defining equivariance and translations, and then move on to studying the existence translational equivariance in different modules.

\paragraph{Translation.} In the context of computer vision, translation typically refers to a shift (and sometimes rotation) in pixels.

\paragraph{Equivariance.} A function $f$ is equivariant to a function $\mathcal{T}$ if $\mathcal{T}(f(x)) = f(\mathcal{T}(x))$.

\paragraph{Translational Equivariance.} An operation $f$ is equivariant to translations.

\paragraph{Linear projections.} A single linear layer, which can also be formulated as a 1\texttimes{}1 convolution, is by definition equivariant to any change in the order of pixels. Therefore, they are also translationally equivariant.

\paragraph{Convolutions.} Thanks to their dynamic sliding window structure, and their static kernel weights, convolutions are translationally equivariant~\cite{goodfellow2016deep}, since every output pixel is the product of its corresponding input pixel centred in a window and multiplied by the static kernel weight.

\paragraph{Self Attention.} SA (\cref{eq:attention}) is translationally equivariant~\cite{ramachandran2019stand}, because: 1. the linear projections maintain that property, and 2. self attention weights are also equivariant to any change in order.

\paragraph{SASA.} SASA~\cite{ramachandran2019stand} extracts key-value pairs for every query according to the same raster-scan pattern convolutions follow, which suggests it maintains translational equivariance.
However, convolutions apply static kernel weights, which allows them to maintain this property.
On the other hand, even though SASA applies dynamic weights, those weights are still a function of the pixels within the window.
Therefore, SASA also maintains translational equivariance.
Note that SASA does not enjoy the same position-agnostic property in self attention.

\paragraph{HaloNet.} The blocked self attention pattern described in HaloNet~\cite{vaswani2021scaling} is described to ``relax'' translational equivariance. 
It is simply due to the fact that pixels within the same region share their neighborhood, therefore their sliding window property is relaxed and with it translational equivariance.

\begin{figure*}[]
    \centering
    \begin{center}
        \setlength{\fboxsep}{0pt} 
        \setlength\tabcolsep{1pt} 
        \renewcommand{\arraystretch}{0.6667} 
            \begin{tabular}{m{2cm}m{2cm}m{2cm}m{2cm}m{2cm}m{2cm}m{2cm}}
                & $x$ & $\mathcal{T}(x)$ & $\text{Sw}(\mathcal{T}(x))$ & $\mathcal{T}(\text{Sw}(x)))$ & $\text{NA}^2(\mathcal{T}(x))$ & $\mathcal{T}(\text{NA}^2(x)))$  \\
                \\
                \textbf{Rotation} & \fbox{\includegraphics[height=2.0cm]{figures/te/orig.png}} & \fbox{\includegraphics[height=2.0cm]{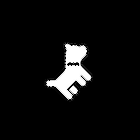}} & \fbox{\includegraphics[height=2.0cm]{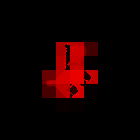}} & \fbox{\includegraphics[height=2.0cm]{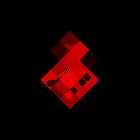}} & \fbox{\includegraphics[height=2.0cm]{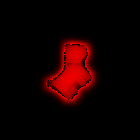}} & \fbox{\includegraphics[height=2.0cm]{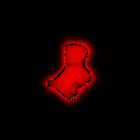}} \\
                \textbf{Shift} & \fbox{\includegraphics[height=2.0cm]{figures/te/orig.png}} & \fbox{\includegraphics[height=2.0cm]{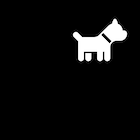}} & \fbox{\includegraphics[height=2.0cm]{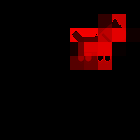}} & \fbox{\includegraphics[height=2.0cm]{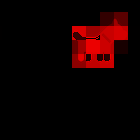}} & \fbox{\includegraphics[height=2.0cm]{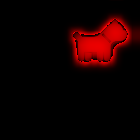}} & \fbox{\includegraphics[height=2.0cm]{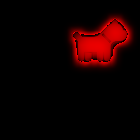}} \\
            \end{tabular}
            \setlength\tabcolsep{6pt} 
        \renewcommand{\arraystretch}{1} 
        \setlength{\fboxsep}{3pt} 
    \end{center}
    \caption{
    \textbf{Visualization of translations applied to Swin and NAT.} 
    $\mathcal{T}$ denotes the translation function (top row is rotation, bottom row is shift).
    ``Sw'' denotes a WSA+SWSA applied to the input, with a residual connection in between.
    This pattern breaks translational equivariance.
    ``NA$^2$'' denotes two layers of NA applied to the input, with a residual connection in between.
    NA preserves translational equivariance with its sliding window property.
    }
    \label{appfig:swinnattr}
\end{figure*}

\paragraph{WSA and SWSA.} The basic property present in both WSA and SWSA is the partitioning, which exists in only one of two forms (regular and shifted) and therefore not dynamically sliding like SASA or convolutions. 
This simply breaks translational equivariance, as translations move the dividing lines.
To give an example, an object within the feature map could fit within a single WSA partition, but the translation could shift the object just enough so that it falls into two different partitions.
To illustrate this, we provide visualizations of activations from a single Swin block (WSA + SWSA) in \cref{appfig:swinnattr}, where we compare translations applied to input and output. 
We replace all linear projections with the identity function (as those are already known to be equivariant) and remove positional biases for simplicity in visualization.

\begin{figure}
    \begin{center}
        \setlength{\fboxsep}{0pt} 
        \setlength\tabcolsep{1pt} 
        \renewcommand{\arraystretch}{0.6667} 
            \begin{tabular}{ccc}
                $x$ & & $\mathcal{T}(x)$  \\
                \\
                \begin{tabular}{l} \fbox{\includegraphics[height=2.0cm]{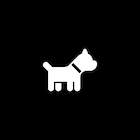}} \end{tabular} & & \begin{tabular}{l} \fbox{\includegraphics[height=2.0cm]{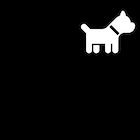}} \end{tabular} \\
                \\
            \end{tabular}
            \begin{tabular}{cccc}
                $f$ & $f(x)$ & $f(\mathcal{T}(x))$ & $\mathcal{T}(f(x)))$  \\
                \\
                \begin{tabular}{l}\textbf{WSA+SWSA}\end{tabular} & \begin{tabular}{l}\fbox{\includegraphics[height=2.0cm]{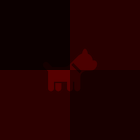}}\end{tabular} & \begin{tabular}{l}\fbox{\includegraphics[height=2.0cm]{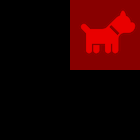}}\end{tabular} & \begin{tabular}{l}\fbox{\includegraphics[height=2.0cm]{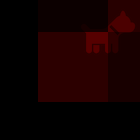}}\end{tabular} \\
                \begin{tabular}{l}\textbf{SASA+SASA}\end{tabular} & \begin{tabular}{l}\fbox{\includegraphics[height=2.0cm]{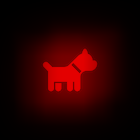}}\end{tabular}  & \begin{tabular}{l}\fbox{\includegraphics[height=2.0cm]{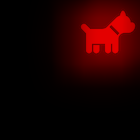}}\end{tabular} & \begin{tabular}{l}\fbox{\includegraphics[height=2.0cm]{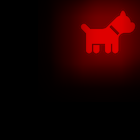}}\end{tabular} \\
                \begin{tabular}{l}\textbf{NA+NA}\end{tabular} & \begin{tabular}{l}\fbox{\includegraphics[height=2.0cm]{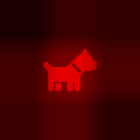}}\end{tabular} & \begin{tabular}{l}\fbox{\includegraphics[height=2.0cm]{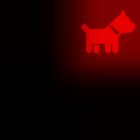}}\end{tabular} & \begin{tabular}{l}\fbox{\includegraphics[height=2.0cm]{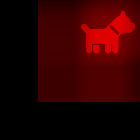}}\end{tabular} \\
            \end{tabular}
            \caption{
            \textbf{Corner pixel visualizations with quarter size kernels.} 
            $x$ denotes the input image with the object centered, $f(x)$ denotes the output when the function $f$ is applied, and $\mathcal{T}$ is the translation that shifts the object to the upper right side corner of the image.
            While SASA does not break translational equivariance at corner pixels as much as NA, it would suffer from a reduced attention span in those areas, which is the reason it does not approach self attention.
            Simply looking at SASA's output for the original centered input shows the effect of the reduced attention span, when compared to NA's output.
            }
            \label{appfig:terelaxation}
            \setlength\tabcolsep{6pt} 
        \renewcommand{\arraystretch}{1} 
        \setlength{\fboxsep}{3pt} 
    \end{center}
\end{figure}

\paragraph{NA.} We note that our NA preserves translational equivariance for the most part, similar to SASA. 
However, NA relaxes translational equivaraince in corner cases in favor of maintaining attention span.
We present translations applied to dummy inputs and their NA outputs in \cref{appfig:swinnattr}, similar to those of Swin.
However, we also note that NA relaxes the translational equivariance in corner cases, particularly because of its definition of neighborhood which results in sliding windows being repeated at edge pixels.
A visualization of this can be seen visualized with a larger kernel size (quarter of the image) compared to Swin and SASA in \cref{appfig:terelaxation}.

The difference in how corner cases are handled is an important difference which should exist between sliding window attention mechanisms and convolutions.
Repeating sliding windows at corner cases (which NA achieves with the neighborhood definition) is useful in the scope of attention, because the repeated windows are still subsets of the original self attention weights, which are being restricted.
This does not hold true in convolutions, where repeated sliding windows produces repeated output pixels, because of the static kernel.
On the other hand, zero padding in attention (no repetition at corner cases; like SASA) is less powerful because it limits attention span farther at corner cases.
It also does not approach self attention as its window size grows, while NA does.

\end{document}